\UseRawInputEncoding
\documentclass[pdflatex,sn-mathphys-num]{sn-jnl}

\usepackage{graphicx}%
\usepackage{multirow}%
\usepackage{amsmath,amssymb,amsfonts}%
\usepackage{amsthm}%
\usepackage{mathrsfs}%
\usepackage[title]{appendix}%
\usepackage{xcolor}%
\usepackage{textcomp}%
\usepackage{manyfoot}%
\usepackage{booktabs}%
\usepackage{algorithm}%
\usepackage{algorithmicx}%
\usepackage{algpseudocode}%
\usepackage{listings}%
\usepackage{textcomp} 
\usepackage{siunitx}
\usepackage{tabularx, booktabs, ragged2e}
\newcolumntype{Y}{>{\RaggedRight\arraybackslash}X} 
\usepackage{array}

\usepackage{mathrsfs}
\usepackage{tabularx} 
\usepackage{booktabs}  
\usepackage{makecell} 
\usepackage{rotating} 
\usepackage{array} 
\usepackage{placeins}
\usepackage{float}
\usepackage{color}
\usepackage{xcolor}
\usepackage{amsmath}

\theoremstyle{thmstyleone}%
\theoremstyle{thmstyletwo}%

\theoremstyle{thmstylethree}%

\begin{document}

\title[Article Title]{Advances in Photoacoustic Imaging Reconstruction and Quantitative Analysis for Biomedical Applications}

\author[1,2]{\fnm{Lei} \sur{Wang}}
\author*[1]{\fnm{Weiming} \sur{Zeng}}\email{zengwm86@163.com}
\author[3]{\fnm{Kai} \sur{Long}}
\author[1]{\fnm{Hongyu} \sur{Chen}}
\author[4]{\fnm{Rongfeng} \sur{Lan}}
\author[3]{\fnm{Li} \sur{Liu}}
\author[5]{\fnm{Wai Ting} \sur{Siok}}
\author*[5]{\fnm{Nizhuan} \sur{Wang}}\email{nizhuan.wang@polyu.edu.hk}

\affil[1]{\orgdiv{The Laboratory of Digital Image and Intelligent Computation}, 
          \orgname{Shanghai Maritime University}, 
          \orgaddress{\postcode{201306}, \city{Shanghai}, \country{China}}}

\affil[2]{\orgdiv{School of Electronic Engineering and Intelligent Manufacturing}, 
          \orgname{Anqing Normal University}, 
          \orgaddress{\postcode{246133}, \city{Anqing}, \country{China}}}

\affil[3]{\orgdiv{School of Engineering, Great Bay University}, 
          \orgname{Dongguan Great Bay Institute for Advanced Study}, 
          \orgaddress{\postcode{523000}, \city{Dongguan}, \country{China}}}

\affil[4]{\orgdiv{Department of Cell Biology \& Medical Genetics}, 
          \orgname{School of Basic Medical Sciences, Shenzhen University Medical School}, 
          \orgaddress{\postcode{518060}, \city{Shenzhen}, \country{China}}}

\affil[5]{\orgdiv{Department of Chinese and Bilingual Studies}, 
          \orgname{The Hong Kong Polytechnic University}, 
          \orgaddress{\postcode{999077}, \city{Hong Kong}, \country{China}}}

\abstract{Photoacoustic imaging (PAI), a modality that combines the high contrast of optical imaging with the deep penetration of ultrasound, is rapidly transitioning from preclinical research to clinical practice. However, its widespread clinical adoption faces challenges such as the inherent trade-off between penetration depth and spatial resolution, along with the demand for faster imaging speeds. This review comprehensively examines the fundamental principles of PAI, focusing on three primary implementations: photoacoustic computed tomography (PACT), photoacoustic microscopy (PAM), and photoacoustic endoscopy (PAE). It critically analyzes their respective advantages and limitations to provide insights into practical applications. The discussion then extends to recent advancements in image reconstruction and artifact suppression, where both conventional and deep learning (DL)-based approaches have been highlighted for their role in enhancing image quality and streamlining workflows. Furthermore, this work explores progress in quantitative PAI, particularly its ability to precisely measure hemoglobin concentration, oxygen saturation, and other physiological biomarkers. Finally, this review outlines emerging trends and future directions, underscoring the transformative potential of DL in shaping the clinical evolution of PAI.}




\maketitle

\section{Introduction}

\subsection{Fundamentals of Photoacoustic Imaging (PAI)}
Optical imaging and ultrasound imaging (US) are well-established modalities that play complementary roles in medical diagnostics \cite{1oraevsky1994laser}. Optical imaging offers high sensitivity and intrinsic molecular contrast but is fundamentally limited by strong optical scattering in biological tissues, which restricts high-resolution imaging to superficial depths \cite{2matthews2017joint}. In contrast, US provides excellent spatial resolution at depths of several centimeters owing to low acoustic scattering, making it indispensable in cardiovascular, abdominal, and obstetric applications \cite{3wang2008tutorial}, \cite{4poudel2019survey}. However, US suffers from low intrinsic contrast because soft tissues exhibit minimal differences in acoustic impedance, and conventional ultrasound contrast agents (e.g., microbubbles) do not enhance optical absorption. This limitation has motivated the development of hybrid imaging approaches that combine the rich optical contrast of absorption-based imaging with deep penetration and high-resolution ultrasound \cite{5ning2015ultrasound}, \cite{6zackrisson2014light}, \cite{7park2025clinical}.

Photoacoustic imaging (PAI) utilizes the photoacoustic effect, a hybrid physical process in which pulsed laser light is absorbed by tissue chromophores, converted into a transient temperature increase, and subsequently generates ultrasonic waves through thermoelastic expansion \cite{7park2025clinical} \cite{8park2017contrast}. As illustrated in Fig.~\ref{fig1}, optical excitation leads to selective absorption by endogenous chromophores---such as oxy- and deoxy-hemoglobin, melanin, lipids, collagen, DNA, and RNA---or by exogenous agents, including clinically approved dyes such as indocyanine green (ICG) and methylene blue. The resulting localized thermal expansion induces broadband acoustic waves that propagate to the tissue surface, where they are detected using an ultrasound transducer array. By analyzing the time-of-flight and amplitude of these signals, a two- or three-dimensional map of the initial optical absorption distribution can be reconstructed, providing high-resolution visualization of both functional and molecular tissue characteristics \cite{7park2025clinical}, \cite{9rajendran2022photoacoustic}.

While small-molecule dyes, such as ICG, offer clinical compatibility, their limited photostability, narrow absorption tunability, and rapid renal clearance restrict their utility in advanced molecular imaging. Engineered nanomaterial-based photoacoustic agents have emerged as promising alternatives to address these limitations. Among them, gold nanostructures (e.g., nanorods and nanocages) leverage localized surface plasmon resonance to achieve strong and spectrally tunable near-infrared absorption, enabling high-contrast imaging and photothermal therapy. However, their clinical translation is hindered by potential long-term accumulation in the reticuloendothelial system (e.g., liver and spleen) and slow biodegradation, raising concerns about chronic toxicity. In contrast, semiconducting polymer nanoparticles  exhibit excellent photostability, high molar extinction coefficients in the NIR-II window, and favorable biodegradability profiles, with several formulations demonstrating efficient renal and hepatobiliary clearance. Perovskite nanocrystals offer ultrahigh absorption coefficients and sharp spectral features; however, they face significant biosafety barriers owing to their lead content, poor colloidal stability in physiological environments, and risk of releasing toxic ions upon degradation. Collectively, the clinical adoption of these nanoplatforms requires rigorous assessment of biocompatibility, clearance kinetics, and regulatory compliance---particularly for agents containing heavy metals or non-biodegradable components \cite{162manohar2011gold}, \cite{163zhang2024exploiting}, \cite{164hsu2023nanomaterial}.

{The efficient generation of photoacoustic signals, irrespective of the absorber type, relies on two key physical conditions: \textit{stress confinement} and \textit{thermal confinement}. Stress confinement requires the laser pulse duration to be shorter than the stress relaxation time (i.e., the time required for the pressure to propagate out of the irradiated volume), ensuring that the induced pressure is not dissipated during laser excitation. Thermal confinement requires negligible heat diffusion during the pulse so that the energy remains localized. Under these conditions, the fractional volume change is governed by the following thermoelastic equation:
\begin{equation}
\frac{\Delta V}{V} = -\kappa \Delta p + \beta \Delta T,
\end{equation}
where $\Delta p$ and $\Delta T$ are the changes in pressure and temperature, respectively; $\kappa$ is the isothermal compressibility; and $\beta$ is the volumetric thermal expansion coefficient \cite{10dispirito2021sounding}. The initial pressure increase $p_0$ is directly proportional to the absorbed optical energy density $E_a = \mu_a F$, yielding
\begin{equation}
p_0 = \frac{\beta}{\kappa \rho c_v} \eta_{\text{th}} \mu_a F = \Gamma \eta_{\text{th}} \mu_a F = \Gamma E_a,
\end{equation}
where $\rho$ is the mass density, $c_v$ is the specific heat capacity at constant volume, $\eta_{\text{th}}$ is the thermal conversion efficiency (typically close to unity for nonradiative absorbers), $\mu_a$ is the absorption coefficient, $F$ is the optical fluence, and $\Gamma = \beta / (\kappa \rho c_v)$ is the dimensionless Grüneisen parameter. This linear relationship forms the basis for quantitative PAI, enabling the estimation of chromophore concentration and oxygen saturation.

\begin{figure*}
\centering
\includegraphics[width=13cm]{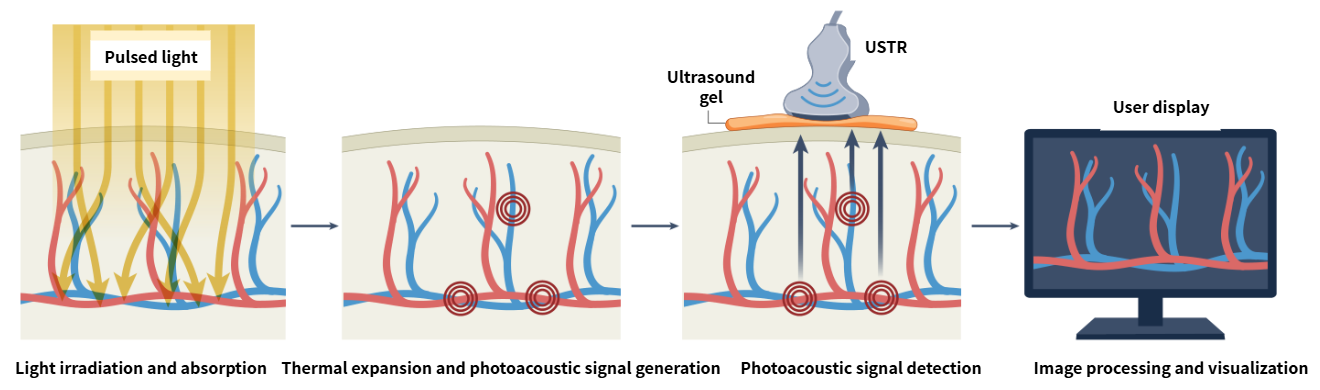}
\caption{Schematic of photoacoustic image generation process \cite{7park2025clinical}.}
\label{fig1}
\end{figure*}

\subsection{Advances of PAI}\label{subsec2_advances}
PAI overcomes the depth limitations of conventional optical imaging by converting absorbed light into detectable ultrasound, thereby enabling high-resolution imaging at depths of approximately \SI{1}{\centi\meter} or greater in scattering tissues \cite{13guo2020photoacoustic}. This hybrid modality facilitates the label-free \textit{in vivo} visualization of morphological, physiological, and metabolic processes \cite{14cox2012quantitative}. For instance, the distinct absorption spectra of oxyhemoglobin and deoxyhemoglobin allow PAI to non-invasively quantify blood oxygen saturation ($\mathrm{sO}_2$), a key functional biomarker in oncology and neuroscience \cite{21liu2019integrated}.

The characteristic performance envelopes of optical coherence tomography (OCT), PAI, and US in terms of their fundamental resolution-depth trade-offs are conceptually illustrated in Fig.~\ref{fig2}. The approximate ranges shown are representative and well-established in the literature: Optical imaging (e.g., OCT) provides high resolution but is limited to superficial depths; US achieves deep penetration at the expense of resolution \cite{180szabo2013diagnostic}, whereas PAI uniquely bridges these two regimes, enabling high-sensitivity optical contrast at ultrasonic scales \cite{179wang2016practical}, \cite{152wang2012photoacoustic}, \cite{18beard2011biomedical}. As summarized in Table~\ref{table1}, PAI uniquely bridges the gap between optical and US; it provides optical-absorption-based contrast superior to US, with a penetration depth far exceeding that of OCT. A key advantage is that PAI does not rely on coherent light detection, thus avoiding inherent coherent speckle noise that degrades OCT images. It is noteworthy that while higher ultrasound frequencies improve the spatial resolution, they concomitantly reduce the imaging depth owing to increased acoustic attenuation \cite{16park2017real}.

Despite challenges such as optical fluence heterogeneity and reconstruction complexity, PAI stands out for its non-invasiveness, absence of ionizing radiation, and molecular specificity, attributes that have driven rapid advances in hardware miniaturization, multispectral unmixing, and deep learning(DL)-based image reconstruction since PAI's first in vivo demonstrations \cite{17mallidi2011photoacoustic,18beard2011biomedical}, with recent advances further exemplified by \cite{19yu2024simultaneous}.

In particular, PAI has emerged as a powerful platform for functional and molecular neuroimaging. Its sensitivity to hemoglobin oxygenation enables noninvasive, label-free mapping of cerebral hemodynamics and $\mathrm{sO}_2$ dynamics during neural activation, making it ideally suited for studying neurovascular coupling, the fundamental link between neuronal activity and the local hemodynamic response. Recent studies have integrated PAI with optogenetic or ultrasonic brain stimulation to probe causal relationships in neural circuits. Advances in high-speed imaging and motion-robust reconstruction now permit functional monitoring in awake, behaving animal models. These developments underscore PAI's growing role in bridging molecular specificity with system-level neuroscience \cite{174zhu2022real}, \cite{175jiang2018estimation}, \cite{176menozzi2025light}.

A prominent example of clinical translation is intravascular photoacoustic imaging (IVPAI). By integrating a miniaturized photoacoustic catheter with intravascular ultrasound), IVPAI enables co-registered structural and compositional imaging of arterial walls. Lipids exhibit strong absorption at characteristic wavelengths (e.g., 1210 and 1720 nm), allowing IVPAI to specifically identify lipid-rich necrotic cores---a hallmark of vulnerable atherosclerotic plaques. Recent clinical studies demonstrated the feasibility of IVPAI in human coronary arteries, whereas hybrid systems combined with multispectral unmixing and DL are paving the way for real-time automated plaque classification \cite{177wang2012vivo}, \cite{178zhou2023multimodal}.

Collectively, by uniquely integrating optical-absorption-based contrast with ultrasound-level penetration depth, PAI is poised to drive transformative advances across a broad spectrum of biomedical applications, from fundamental neuroscience and cancer research to clinical cardiology and real-time intraoperative guidance.

\begin{table}[htbp]
\centering
\caption{Comparative analysis of OCT, US, and PAI. For PAI, values are given as PAM/PACT and represent typical performance.}
\label{table1}
\renewcommand{\arraystretch}{1.4}
\setlength{\tabcolsep}{8pt}
\begin{tabular}{
  >{\bfseries\raggedright\arraybackslash}p{2.8cm}
  >{\raggedright\arraybackslash}p{1.7cm}
  >{\raggedright\arraybackslash}p{1.7cm}
  >{\raggedright\arraybackslash}p{4.6cm}
}
\toprule
\textbf{Characteristic} & \textbf{OCT} & \textbf{US} & \textbf{PAI (PAM/PACT)} \\
\midrule
Imaging depth (mm) & 1--2 & 20--100 & 0.1--1 (PAM) / 10--80 (PACT) \\
Axial resolution (\si{\micro\meter}) & 5--15 & 100--500 & 1--15 (PAM) / 50--200 (PACT) \\
Lateral resolution (\si{\micro\meter}) & 5--20 & 200--1000 & 0.5--5 (PAM) / 200--1000 (PACT) \\
Contrast mechanism & Optical scattering & Acoustic impedance & Optical absorption (hemoglobin, melanin, lipids, contrast agents) \\
Signal texture & Coherent speckle & Acoustic speckle & Minimal in well-defined absorbers such as blood vessels;  texture-like patterns may appear in heterogeneous or scattering tissues. \\
Functional contrast & Limited (angiography) & Moderate (Doppler, elastography) & High (sO\textsubscript{2}, total Hb, flow, metabolism) \\
Clinical adoption & Established (ophthalmology) & Widespread & Emerging (oncology, neurology, cardiovascular, dermatology) \\
Representative applications & Retinal imaging, intravascular coronary imaging & Abdominal organ imaging, fetal monitoring & Oncology (breast cancer, melanoma), Neurology (functional brain imaging), Cardiovascular (atherosclerotic plaque detection), Dermatology (microvasculature mapping) \\
\bottomrule
\end{tabular}
\smallskip
\small
\noindent\textit{Note: Values are approximate and depend on system configuration (e.g., transducer frequency, laser parameters).}
\end{table}

\section{Key Implementations of PAI}
The study of the photoacoustic effect, spurred by advancements in laser technology from the late 1970s to the early 1980s, laid the foundation for PAI \cite{21liu2019integrated}. This groundwork culminated in the early 21st century with the maturation development of key implementations, such as photoacoustic compute tomography (PACT), photoacoustic microscopy (PAM), and photoacoustic endoscopy (PAE)  , which significantly advanced the field and expanded its biomedical applications, opening new avenues for research and clinical practice \cite{22attia2019review}, \cite{23omar2019optoacoustic}, \cite{24moothanchery2017performance}. Fig.~\ref{fig3} illustrates the three main methodologies employed in PAI.

\begin{figure}[htp]
    \centering
    \includegraphics[width=0.8\textwidth]{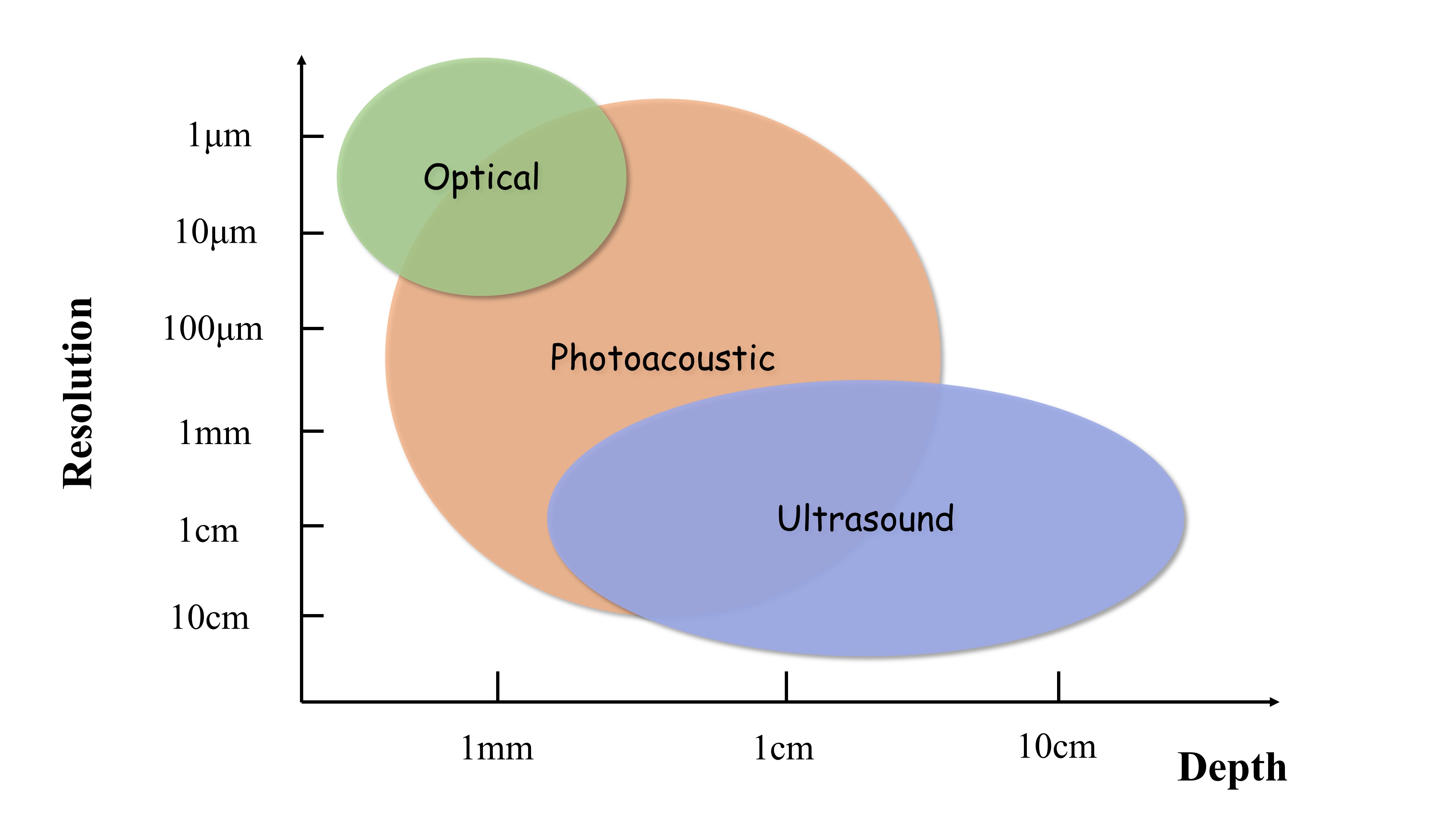}
    \caption{Bridging the imaging gap: comparison of resolution versus depth in photoacoustic, ultrasound, and optical imaging.}
    \label{fig2}
\end{figure}

\begin{figure}[]
    \centering
    \includegraphics[width=13cm]{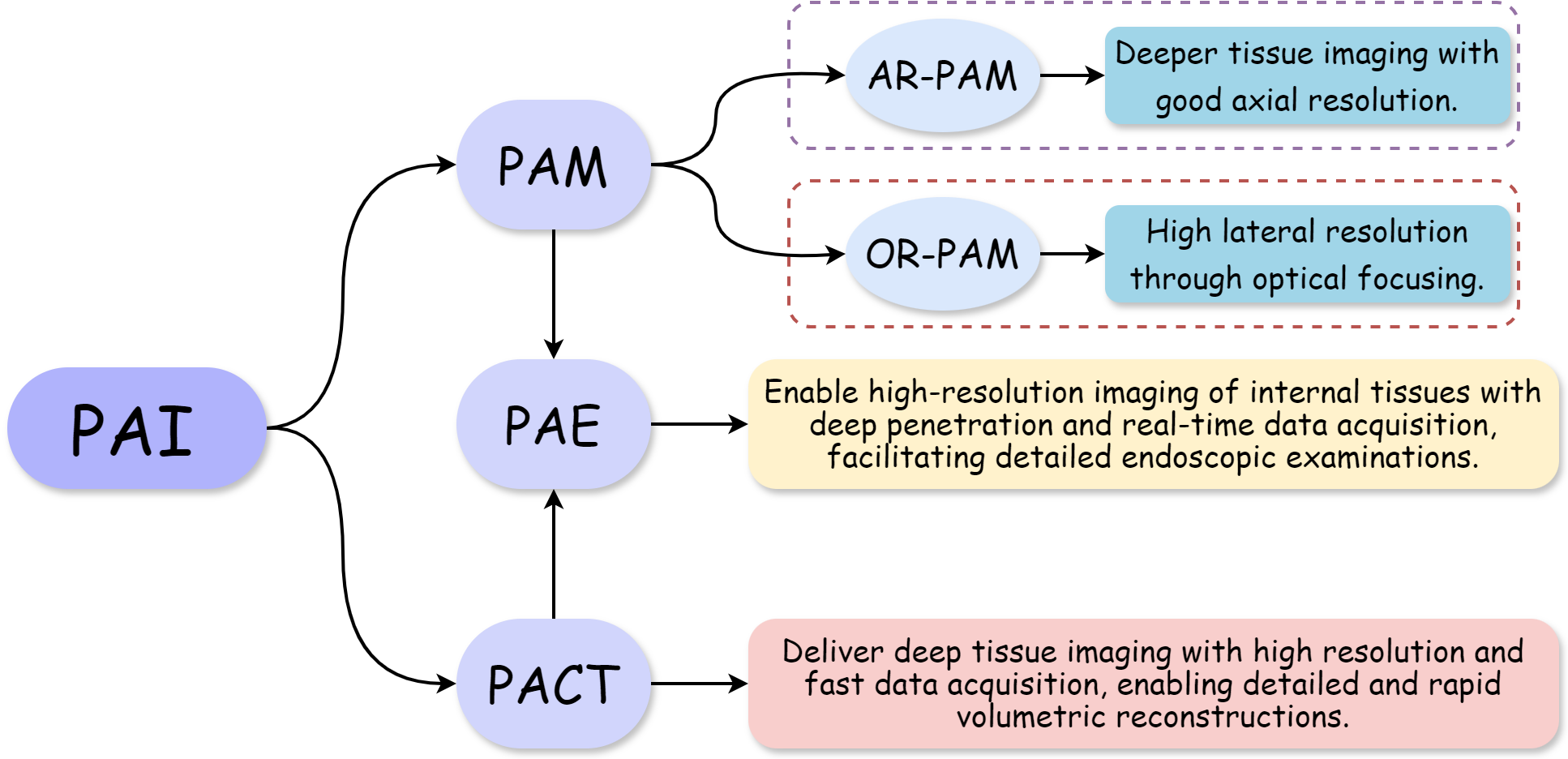}
    \caption{Three key implementations of PAI.}
    \label{fig3}
\end{figure}

\subsection{PACT}
PACT is a rapidly advancing imaging modality that leverages the photoacoustic effect to provide high-contrast, non-ionizing imaging of biological tissues \cite{26_2021Recent} \cite{27na2021photoacoustic}. It offers a wide field of view and deep penetration, enabling diverse biomedical applications such as whole-organ imaging, vascular mapping, and quantitative monitoring of blood oxygen saturation ($\mathrm{sO}_2$) in cancer and neuroscience \cite{14cox2012quantitative}, \cite{28bauer2011quantitative}.

In PACT, a wide-diameter unfocused pulsed laser beam illuminates the tissue surface, and the resulting photoacoustic signals are captured by an array of ultrasonic detectors \cite{22attia2019review}. Inversion algorithms process these signals to reconstruct the initial acoustic field, yielding high-resolution images at scales from the cellular to the organ level, with details down to hundreds of micrometers \cite{29tian2021spatial}, \cite{30jeon2013multimodal}. PACT offers moderate spatial resolution, significant penetration depth, and fast imaging speed. The selection of an ultrasonic transducer array depends on the specific application. Linear arrays are suitable for small animals or surface tissues, whereas hemispherical arrays are better suited for broader human imaging, particularly for breast cancer detection \cite{31jeon2019review}, \cite{32vu2020generative}. The potential use of PACT in breast cancer imaging has driven significant advancements in clinical research, leading to the development of multiple imaging systems \cite{33steinberg2019photoacoustic}. 
Despite its advantages in deep-tissue imaging, the spatial resolution of PACT in biological tissues typically ranges from 50 to 150 \si{\micro\ meter}. This resolution is lower than that of OCT but comparable to or slightly finer than that of clinical ultrasound \cite{34rajendran2020deep}. In addition, the contrast mechanisms differ between the macroscopic and microscopic imaging scales \cite{17mallidi2011photoacoustic}. Table~\ref{table2} summarizes PACT's advantages and constraints in comparison to US, OCT, and conventional optical imaging modalities.

\begin{table}[t]
\centering
\caption{Advantages and constraints of PACT modality in comparison with US, OCT, and conventional optical imaging techniques.}
\label{table2}
\renewcommand{\arraystretch}{1.5}
\setlength{\tabcolsep}{6pt}
\begin{tabularx}{\textwidth}{m{1.5cm} X X}
\toprule
\textbf{Modality} & \textbf{Advantages} & \textbf{Constraints} \\ 
\midrule
PACT &  
\begin{tabular}[t]{@{}l@{}} 
Non-invasive and \\free from ionizing radiation \\
Wide field of view \\
Greater imaging depth than OCT and \\conventional optical imaging, enabling \\millimeter-to-centimeter scale imaging\\
Fast imaging speed\\(enabling real-time or functional imaging) \\
Spatial resolution between that of US and \\ OCT (typically 50--200\,\si{\micro\meter}) \\
Versatile across cellular-to-organ scales
\end{tabular} &  
\begin{tabular}[t]{@{}l@{}} 
Higher system cost compared to US \\
Lower spatial resolution than OCT \\and high-frequency ultrasound \\
Lacks subcellular resolution, \\limiting integration with high-resolution \\optical microscopy
\end{tabular} \\  
\bottomrule
\end{tabularx}
\end{table}

Recent advancements in PACT are driven by innovations in imaging devices, contrast agents, and models \cite{35zhang2024challenges}. PACT relies on reconstruction algorithms to transform the detected photoacoustic signals into acoustic source distributions, thereby generating high-resolution images \cite{36tian2020impact}. These algorithms play crucial roles in determining imaging accuracy and quality \cite{37yao2021perspective}. However, several challenges remain, including limited angular coverage and constrained detection bandwidth, both of which can degrade image resolution and contrast \cite{36tian2020impact}, \cite{38zhou2016tutorial}. The reconstruction process inherently involves solving an ill-posed inverse acoustic problem (IAP) using incomplete data. To further enhance PACT's utility in biomedical imaging, it is essential to optimize existing technologies and develop novel methodologies.

These technological and algorithmic advancements have paved the way for DL-enhanced PACT. Recent studies have demonstrated the potential of longitudinal tumor monitoring and therapy response evaluation. By leveraging PAI's sensitivity to oxygenation and vascular dynamics, DL-based reconstruction enables the visualization of temporal changes in tumor hypoxia, angiogenesis, and metabolic reprogramming during immunotherapy and photothermal therapy. Furthermore, data-driven quantitative frameworks improve temporal consistency and allow predictive modeling of treatment efficacy, thereby highlighting the translational promise of PAI for noninvasive cancer therapy assessment \cite{173langley2025heterogeneous}.

\begin{sidewaystable}
\centering
\caption{Comparative analysis of OR-PAM and AR-PAM modalities.}
\label{table3}
\renewcommand{\arraystretch}{1.5}
\begin{tabular}{@{}>{\raggedright\arraybackslash}p{3.8cm}
                  >{\raggedright\arraybackslash}p{4.6cm}
                  >{\raggedright\arraybackslash}p{4.6cm}@{}}
\toprule
\textbf{Characteristics} & \textbf{OR-PAM} & \textbf{AR-PAM} \\
\midrule

\textbf{Advantages} & & \\
\midrule

Lateral Resolution & High ($<5\,\mathrm{\mu m}$) & Moderate ($>50\,\mathrm{\mu m}$) \\

Axial Resolution & Determined by ultrasound transducer bandwidth & Determined by ultrasound transducer bandwidth \\

Depth Penetration & Limited ($\sim$1--2\,mm) & Deeper ($\sim$3--10\,mm) \\

Contrast Mechanism & Enables high-resolution visualization of microvasculature and cellular structures & Suitable for functional imaging beyond 1\,mm depth \\

Imaging Precision & Suitable for high-precision surface imaging & Suitable for deeper tissue imaging \\

\midrule
\textbf{Constraints} & & \\
\midrule

Depth Penetration & Restricted by light scattering effects & Hindered by low resolution and high background noise \\

Lateral Resolution & Limited applicability for deep tissue imaging & Lower resolution due to acoustic focusing \\

Imaging Speed & Speed constrained by mechanical scanning & Slower due to broader imaging coverage \\

Contrast Noise Ratio & Challenged by reduced contrast in deeper tissues & Higher background noise impacting clarity \\

Clinical Application & Constrained by depth and speed limitations & Challenged by lower resolution and noise in deeper tissues \\

\bottomrule
\end{tabular}
\end{sidewaystable}

\subsection{PAM}
Over the past decade, PAM has emerged as a crucial technique for microscopic imaging owing to its distinctive optical absorption contrast mechanism \cite{27na2021photoacoustic} \cite{30jeon2013multimodal}. As a versatile biomedical imaging modality, PAM facilitates structural, functional, and molecular imaging by using both endogenous and exogenous contrast agents. It has been widely applied in blood flow perfusion, oxygenation imaging, tumor visualization, and neuroimaging by leveraging hemoglobin as an intrinsic optical absorber \cite{38zhou2016tutorial}, \cite{40yao2013photoacoustic}.

PAM is broadly categorized into optical-resolution PAM (OR-PAM) and acoustic-resolution PAM (AR-PAM) depending on whether the lateral resolution is governed by the optical or acoustic focus \cite{5ning2015ultrasound}, \cite{18beard2011biomedical}, \cite{25lin2022emerging}.

While PACT achieves centimeter-scale penetration depths, PAM---particularly OR-PAM---delivers a significantly higher spatial resolution (typically $<\SI{5}{\micro\meter}$) in superficial tissues (\SIrange{1}{2}{\milli\meter}), and a depth range accessible to both modalities. However, PAM has been optimized for surface and near-surface imaging \cite{18beard2011biomedical}, \cite{33steinberg2019photoacoustic}.

In OR-PAM, a tightly focused laser enables micron-scale lateral resolution but is limited to depths of \SIrange{1}{2}{\milli\meter} owing to optical scattering. 
In contrast, AR-PAM uses weakly focused or collimated illumination, which shifts the resolution limit to the acoustic focus (typically $>\SI{50}{\micro\meter}$) and enables greater imaging depths of \SIrange{3}{10}{\milli\meter}. However, a larger illuminated area can result in stronger background signals. In both configurations, the axial resolution is determined by the bandwidth of the ultrasound transducer. A key challenge in AR-PAM is improving the lateral resolution without sacrificing penetration depth.

Most OR-PAM systems rely on mechanical scanning for both optical excitation and ultrasound detection, which limits their imaging speeds. Nevertheless, OR-PAM remains invaluable for high-resolution biomedical applications, such as tumor angiography and melanoma cell imaging \cite{31jeon2019review}, \cite{41kempski2020application}. AR-PAM, which overcomes the optical diffusion limit, is well suited for deeper functional imaging beyond 1\,\si{\milli\meter}, enabling the quantification of oxygenated and deoxygenated hemoglobin concentrations, blood flow velocity, and oxygen metabolism rates \cite{20wang2009multiscale}.

One of the primary technical challenges in PAM is the optimization of the tradeoff between spatial resolution and imaging speed \cite{41kempski2020application}, \cite{42chen2021progress}. Despite its potential, the clinical translation of PAM has been constrained by several interdependent challenges, including high laser excitation doses, limited imaging throughput, and difficulties in consistently achieving optimal image quality \cite{17mallidi2011photoacoustic} \cite{31jeon2019review}. Therefore, current research efforts are focused on developing advanced image and signal processing techniques to address these limitations \cite{43hsu2021comparing}. Table~\ref{table3} summarizes the advantages and constraints of PAM.

\begin{table}[htbp]
\centering
\caption{Advantages and constraints of PAE modality.}
\label{table4}
\renewcommand{\arraystretch}{1.3}
\setlength{\tabcolsep}{8pt}
\begin{tabularx}{\textwidth}{>{\raggedright\arraybackslash}p{1.8cm} X X}
\toprule
\textbf{Characteristics} & \textbf{Advantages} & \textbf{Constraints} \\
\midrule
\multirow{5}{*}{PAE} & 
High optical contrast for superficial tissues &
Limited lateral resolution (typically \SIrange{50}{200}{\micro\meter}) \\
& 
Functional imaging capability (e.g., oxygen saturation, hemoglobin concentration) &
Restricted field of view due to lumen geometry \\
& 
Compatibility with endoscopic platforms for \textit{in vivo} use &
Susceptibility to physiological motion artifacts (e.g., peristalsis, respiration) \\
& 
Potential for multimodal integration (e.g., with OCT or fluorescence) &
Signal attenuation in highly scattering or absorbing luminal tissues \\
& 
Miniaturized probe design enables access to confined anatomical sites &
Challenges in system miniaturization and signal-to-noise ratio optimization \\
\bottomrule
\end{tabularx}
\end{table}

\subsection{PAE}
Conventional PACT and PAM are unsuitable for imaging internal structures such as the digestive tract and vascular walls. To overcome this limitation, researchers integrated photoacoustic sensing with endoscopic technology, giving rise to PAE for accessing deeper tissues \cite{19yu2024simultaneous}, \cite{44yoon2013recent}. Although PAE has emerged more recently than PACT and PAM, it has rapidly advanced and can image internal tissues, such as the digestive tract, blood vessels, and the urogenital system, emerging as a significant area of research \cite{44yoon2013recent}.

PAE is particularly suitable for detecting lesions in biological cavities, such as the nasal cavity, digestive tract, and arterial vessels, and offers high-quality images of superficial tissues for diagnostic purposes. It utilizes miniaturized MEMS scanning devices and fiber-optic sensors to reduce the system size, enabling in vivo imaging for the examination of organ conditions or lesions. However, spatial constraints within the lumen limit the angular range of the detector for collecting photoacoustic signals, which may compromise the image quality. Table \ref{table4} summarizes the advantages and constraints of PAE, and Table \ref{table5} compares the characteristics of the three principal PAI modalities.

Recently, PAE has shown remarkable potential for both preclinical and translational biomedical applications \cite{13guo2020photoacoustic}, \cite{143zhang2025recent}, \cite{146xu2006photoacoustic}. In gastrointestinal imaging, miniaturized photoacoustic endoscopes have been developed to map vascular morphology, oxygen saturation, and inflammatory responses in organs such as the esophagus, stomach, and colon \cite{13guo2020photoacoustic}, \cite{143zhang2025recent}. These systems enable the early detection of precancerous and neoplastic lesions by resolving microvascular networks with high optical contrast. In cardiovascular research, intravascular PAE systems have been utilized to characterize atherosclerotic plaques, lipid distribution, and vascular wall composition, offering morphological and molecular contrasts that complement traditional ultrasound and OCT. Moreover, prototype PAE probes have been applied to the urinary and reproductive tracts for the real-time visualization of mucosal structures and microvasculature, highlighting their potential for minimally invasive diagnosis and intraoperative monitoring of urogenital diseases \cite{144yang2024perspectives}, \cite{143zhang2025recent}. Collectively, these studies underscore the versatility of PAE as a cross-disciplinary imaging platform capable of providing structural and functional insights \cite{13guo2020photoacoustic}, \cite{146xu2006photoacoustic}.%

Despite rapid progress, several challenges remain before PAE can achieve widespread clinical deployment \cite{13guo2020photoacoustic}, \cite{143zhang2025recent}. Miniaturization and system integration are constrained by the narrow diameter of biological lumens, which necessitates delicate trade-offs between spatial resolution, imaging depth, and field of view \cite{142sun2025dual}, \cite{143zhang2025recent}. In vivo imaging stability is further affected by physiological motions such as respiration and peristalsis, whereas limited-view detection geometries can lead to signal dropout and reconstruction artifacts. Overcoming these issues requires continued innovation in compact scanning mechanisms, flexible probe architectures, and computational reconstruction algorithms capable of compensating for incomplete acoustic sampling \cite{145shang2017simultaneous}, \cite{144yang2024perspectives}, \cite{146xu2006photoacoustic}. Emerging trends include the development of multimodal endoscopic systems that integrate PAE with OCT, fluorescence, or US, as well as DL–based frameworks for artifact correction and real-time reconstruction \cite{141liang2022optical}, \cite{144yang2024perspectives}, \cite{146xu2006photoacoustic}. With continued advances in MEMS technology, fiber-optic detection, and intelligent reconstruction algorithms, PAE is expected to evolve into a clinically viable tool for early diagnosis, functional assessment, and intraoperative guidance in gastrointestinal, vascular, and urogenital applications \cite{13guo2020photoacoustic} \cite{143zhang2025recent}.

The subsequent section delves into image reconstruction techniques for PAI, covering both conventional algorithms and deep-learning-based frameworks, and provides a quantitative analysis of PAI, as outlined in Fig. \ref{fig4}.

\begin{table}[htbp]
\centering
\caption{Comparison of characteristics among three PAI modalities.}
\label{table5}
\renewcommand{\arraystretch}{1.5} 
\setlength{\tabcolsep}{8pt} 
\begin{tabular}{>{\raggedright\arraybackslash}p{2cm} 
                >{\raggedright\arraybackslash}p{1.5cm} 
                >{\raggedright\arraybackslash}p{3.5cm} 
                >{\raggedright\arraybackslash}p{1.5cm} 
                >{\raggedright\arraybackslash}p{1.5cm}}
\toprule
\textbf{Modalities} & \textbf{Resolution} & \textbf{Field of View} & \textbf{Depth} & \textbf{Cost} \\
\midrule
PACT       & Low  & Large                     & Deep     & Expensive \\
AR-PAM     & Moderate & Limited by scanning range & Moderate & Moderate   \\
OR-PAM     & High & Small                     & Shallow  & Expensive \\
PAE        & Moderate & Small                     & Shallow  & Moderate   \\
\bottomrule
\end{tabular}

\smallskip
\small
\noindent\textit{Note:} Resolution rankings are relative among PAI modalities. ``Shallow'' refers to depths $<2$\,mm (e.g., skin, epidermis); ``Moderate'' refers to $2$--$10$\,mm (e.g., small animal organs); ``Deep'' refers to $>10$\,mm up to several centimeters (e.g., breast, brain). Cost categories are qualitative: ``Expensive'' indicates systems requiring ultrafast lasers and high-end detection components (e.g., PACT, OR-PAM); ``Moderate'' refers to systems using standard or lower-cost components (e.g., AR-PAM, PAE).
\end{table}

\begin{figure}[]
    \centering
    \includegraphics[width=0.75\textwidth]{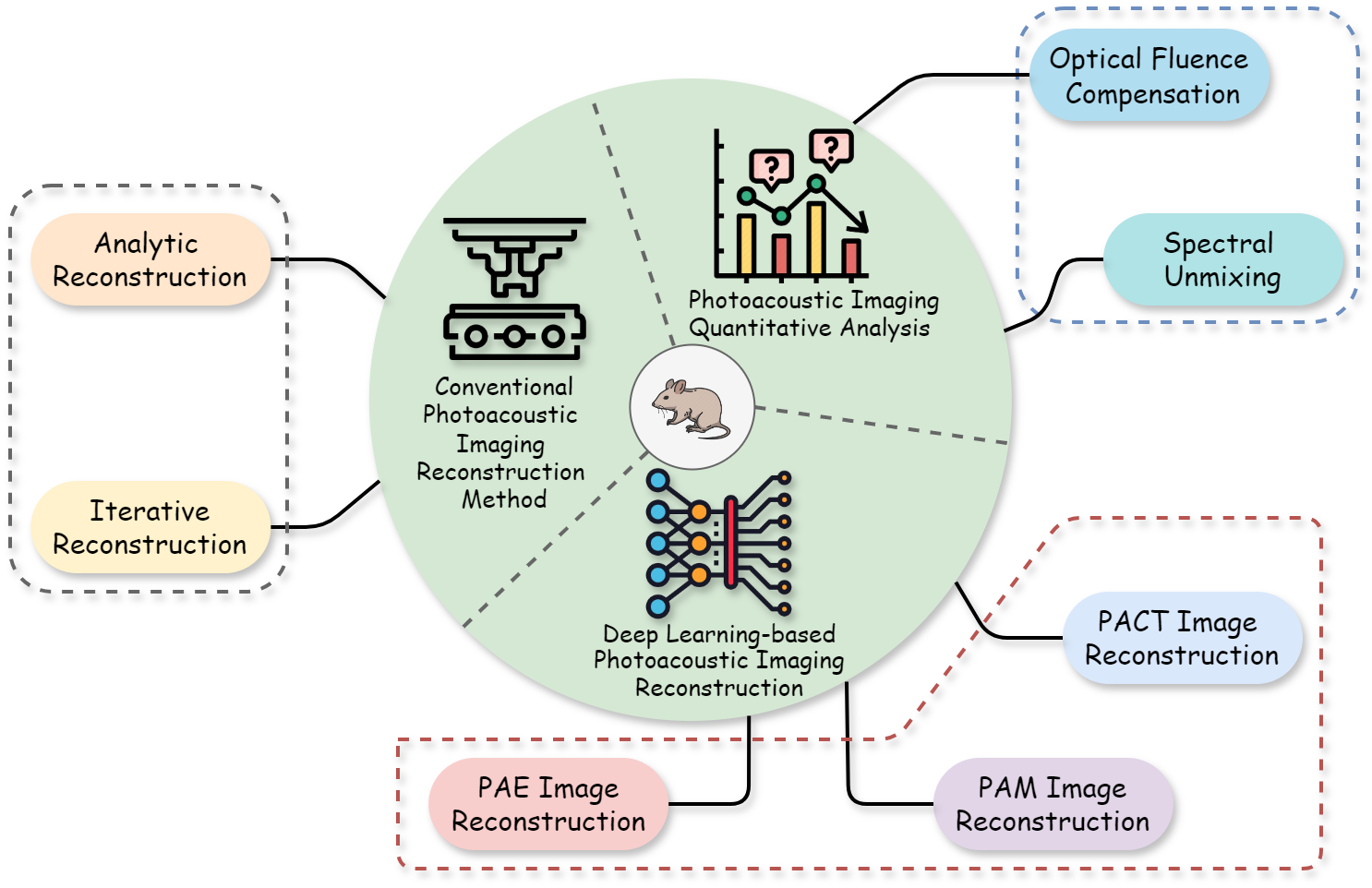}
    \caption{Illustrative overview outlining the scope of this review.}
    \label{fig4}
\end{figure}

\begin{figure}[]
    \centering
    \includegraphics[width=13cm]{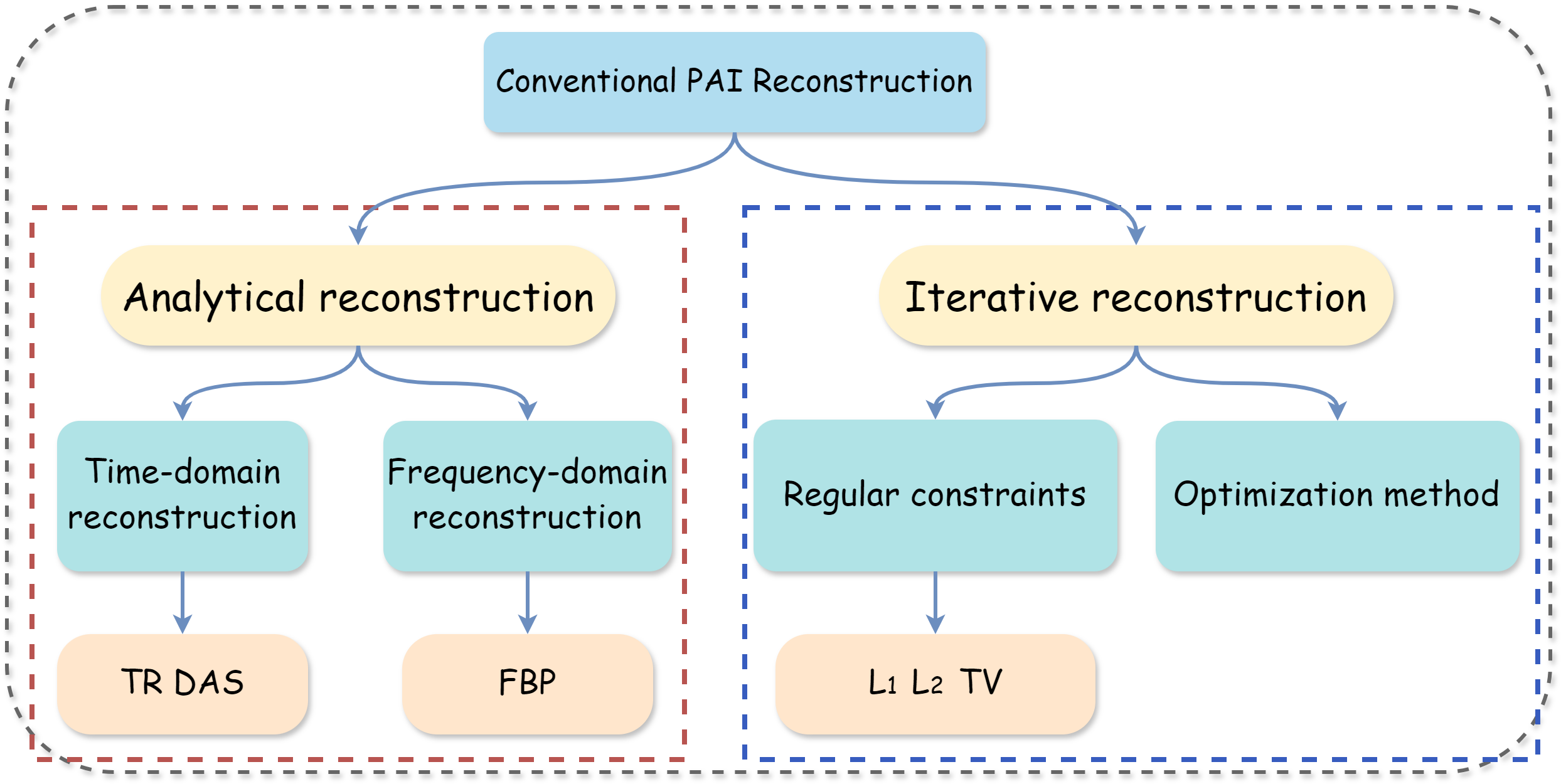}
    \caption{Illustration of conventional PAI reconstruction methods.}
    \label{fig5}
\end{figure}

\section{Conventional Reconstruction in PAI}
The core principle of PAI is to map the optical absorption coefficient distribution within the tissues, which is directly correlated with the initial acoustic pressure distribution. Developing robust reconstruction models is crucial for accurately reconstructing the initial acoustic pressure from acquired photoacoustic signals \cite{45Seong2020}.

Conventional reconstruction methods can be broadly classified into analytical and iterative methods \cite{46_2024Image} as shown in Fig. \ref{fig5}. Analytical reconstruction methods use time- and frequency-domain techniques to find optimal solutions, such as time reversal (TR) \cite{47xu2004time}, \cite{48burgholzer2007exact}, delay-and-sum (DAS) \cite{49hoelen1998three}, \cite{50hoelen1998photoacoustic} in the time domain, and filtered back projection (FBP) \cite{51kruger1995photoacoustic} in the frequency domain.

High-fidelity image reconstruction is vital for improving the accuracy of PAI because it converts raw signals from ultrasonic transducers into images of the initial pressure distributions \cite{52guo2016deep}. When data acquisition is incomplete, conventional PAI reconstruction methods such as back projection \cite{53xu2005universal}, TR, and DAS may result in diminished image quality and depth \cite{54tai2009se}. 

Unlike analytical methods, iterative reconstruction uses optimization and regularization to solve equations, offering higher signal-to-noise ratios (SNR) and improved image quality, but it requires more computational power. It assumes ideal conditions such as high sampling rates and uniform sound speed, which are often not met in real scenarios, leading to errors and degraded image quality. Although iterative methods can partially mitigate these issues, they require substantial computational resources and careful selection of regularization techniques \cite{55wang2023photoacoustic}. 

Despite significant engineering efforts to address the technical challenges of PAI, most solutions rely on complex, expensive hardware and time-consuming image reconstruction processes, including iterative methods. These approaches often require balancing parameters, such as speed vs. field of view and resolution vs. penetration depth. Therefore, innovative approaches that address these challenges from a new perspective are required \cite{56shan2019simultaneous}.

\section{DL-based Reconstruction in PAI}
DL has advanced rapidly, driven by large datasets and powerful computing resources, and is now crucial for tasks such as image classification, segmentation, reconstruction, superresolution, and disease prediction. PAI reconstruction surpasses traditional algorithms by generating high-quality images with enhanced SNR, even at low pulse energy levels \cite{57ravishankar2019image}. Its integration into PAI significantly improves artifact removal, denoising, and super-resolution \cite{32vu2020generative}, \cite{51kruger1995photoacoustic}, \cite{54tai2009se}, and enhances image quality under non-ideal conditions, addressing challenges such as sparse sampling, limited views, restricted bandwidth, heterogeneous media, finite aperture sizes, and insufficient laser power \cite{58davoudi2019deep}.  

Antholzer\textit{ et al}.  \cite{59antholzer2018photoacoustic} introduced CNNs for PAI reconstruction in 2017, enhancing reconstruction speed and image quality by learning complex mappings and optimal features. In PACT, DL models yield high-fidelity 3D images \cite{60wei2024deep}, and in PAM, they enable high-contrast, high-resolution imaging of biological tissues \cite{87cheng2022high}; and in PAE, they improve clarity and detail by learning rich features from limited datasets \cite{13guo2020photoacoustic}.

More recently, the landscape of DL in PAI has been reshaped by generative foundation models and self-supervised paradigms, emerging from 2023 to 2025. 
Recent advances in deep generative models have opened new frontiers in PAI reconstruction. 
For instance, Song~\textit{et~al.}~\cite{149song2023sparse} leveraged a diffusion-based prior within a model-based iterative framework to achieve high-quality PAT reconstruction from as few as 32 projections, reporting structural similarity index measure (SSIM) and peak signal-to-noise ratio (PSNR) improvements of 0.65 and 5.1~dB over delay-and-sum, respectively. 
Lian \textit{et al.}~\cite{151lian2025generative} introduced a generative prior constrained accelerated iterative reconstruction method that significantly accelerated PAM by reconstructing high-quality images from undersampled data, thereby achieving a single reconstruction time of approximately 5 s.
Furthermore, Song \textit{et al.}~\cite{150song2024multiple} proposed a framework utilizing multiple diffusion models for extremely limited-view reconstruction, which integrates multiscale diffusion models to mitigate brightness distortion and structural blurring, significantly enhancing image quality even under severe angular limitations (e.g., 60° view).

Moreover, the rise of vision foundation models inspired the development of \textit{training-free} adaptation strategies for PAI. 
Deng \textit{et al.}~\cite{147deng2025streamlined} recently demonstrated that off-the-shelf foundation models (e.g., segment anything model(SAM) can be directly applied to photoacoustic image processing tasks such as segmentation and enhancement without any task-specific fine-tuning, providing a practical solution to the scarcity of labeled clinical data.

Simultaneously, self-supervised learning frameworks are also gaining momentum. 
Lan \textit{et al.}~\cite{148lan2024masked} introduced a masked cross-domain self-supervised framework for PACT in which the network learns robust representations by reconstructing masked regions across simulated and experimental domains, thereby eliminating the need for ground-truth images.

Collectively, these emerging approaches leverage the generative power of diffusion models, the generalizability of foundation models, and data efficiency of self-supervised learning to produce physically plausible solutions. This significantly enhances model generalizability while reducing the critical dependence on large, fully annotated datasets.

Despite these promising advances, each paradigm has distinct trade-offs. Diffusion models can achieve high image fidelity even with limited training data but suffer from slow sampling speeds and high computational demands, limiting their use in real-time applications. In contrast, the training-free adaptation of foundation models offers remarkable flexibility and zero-shot capability, yet may struggle with domain-specific features unique to photoacoustic signals, such as wavelength-dependent absorption and acoustic dispersion, potentially compromising quantitative accuracy. While alleviating the need for paired data, self-supervised approaches often rely on carefully designed pretext tasks or simulation-to-real transfer assumptions, which may not be generalizable across heterogeneous tissue types or imaging systems. Crucially, none of these methods currently provide uncertainty quantification or guarantee physical consistency, which are key requirements for clinical deployment. Thus, the choice of reconstruction strategy must be guided not only by image quality metrics but also by practical constraints such as speed, data availability, interpretability, and clinical validation requirements.

\subsection{PACT Image Reconstruction}
PACT is a well-established preclinical technique that is gaining traction in clinical translation, reconstructing initial pressure distributions from acoustic signals to enable high-contrast, deep-tissue visualization \cite{36tian2020impact}. A key challenge is achieving high SNR images using cost-effective equipment. Sparse sampling often results in a poor reconstruction using artifacts \cite{39balci2025enhanced} \cite{42chen2021progress}. DL methods have been developed and applied in PACT to enhance the image reconstruction quality, as shown in Tables \ref{table6} \cite{49hoelen1998three}, \cite{50hoelen1998photoacoustic}, \cite{61yang2021review}, \cite{62huang2023review}, \cite{63schmidhuber2015deep}, \cite{64ronneberger2015u}, \cite{65allman2018photoacoustic}. Figure \ref{fig6} illustrates the integration of acoustic inversion and DL in processing radio frequency data for image reconstruction, highlighting the preprocessing, postprocessing, and direct processing approaches \cite{10dispirito2021sounding}. 

\FloatBarrier
\begin{table}[htbp]
\centering
\caption{Brief summary of DL-based reconstruction approaches for PACT.}
\label{table6}
\renewcommand{\arraystretch}{1.5}
\setlength{\tabcolsep}{6pt}
\footnotesize
\begin{tabularx}{\textwidth}{
    >{\raggedright\arraybackslash}m{2.6cm}
    >{\raggedright\arraybackslash}m{1.6cm}
    >{\raggedright\arraybackslash}X
    >{\raggedright\arraybackslash}X
}
    \toprule
    \textbf{DL Methods} & \textbf{Models} & \textbf{Backbones} & \textbf{Key Achievements} \\
    \midrule

    DL-based reconstruction during Pre-processing Stage
    & FC-DNN \cite{66gutta2017deep} 
    & A simple fully connected DNN that enhances bandwidth by mapping band-limited PA signals to full-bandwidth estimates.
    & On PAT phantom: achieves PC = 0.75 / CNR = 2.54 vs. 0.22/0.01 (limited BW) and 0.32/0.29 (Least square deconvolution). \\

    \midrule

    DL-based reconstruction within Direct-processing Stage
    & ResUnet \cite{69feng2020end} 
    & End-to-end ResUnet with residual blocks for direct inversion of PA signals to initial pressure distribution.
    & On numerical and physical phantoms: achieves $>95\%$ higher Pearson correlation and $+39\%$ PSNR vs. MRR, and $+18\%$ PSNR over UNet++ in simulations. \\

    \midrule

    \multirow{5}{2.6cm}{DL-based reconstruction throughout Post-processing Stage}
    & FD-UNet \cite{70guan2019fully} 
    & Fully dense UNet with enhanced feature transmission and reduced redundant learning for sparse PAT image artifact removal.
    & On mouse brain vasculature dataset: achieves 21.12 ± 1.18 PSNR and 0.65 ± 0.04 SSIM (15 detectors), outperforming UNet by +0.91 PSNR and +0.05 SSIM; gains up to +3.37 PSNR and +0.13 SSIM at 45 detectors. \\
    \cmidrule{2-4}
    
    & PAT-AND \cite{74zhong2024unsupervised} 
    & Unsupervised artifact disentanglement network (ADN) with content- and artifact-specific encoders/decoders for unpaired image-to-image translation in sparse-view PAT. 
    & On in vivo data under 16 projections: improves SSIM by $\sim$188\% and PSNR by $\sim$85\% over traditional reconstruction methods. \\
    \cmidrule{2-4}
    
    & MDAEP \cite{75song2024accelerated} 
    & Model-based iterative reconstruction accelerated by multi-channel denoising autoencoder priors for sparse-view PAT.
    & On in vivo data with 32 projections: achieves $+48\%$ PSNR and $+12\%$ SSIM over U-Net, with significantly accelerated convergence. \\
    \cmidrule{2-4}
    
    & DuDoUNet \cite{71zhang2021limited} 
    & Dual-domain U-Net with Information Sharing Block (ISB) and mutual information (MI) prior for limited-view PAT reconstruction using time- and k-space inputs. 
    & On a public clinical dataset: achieves SSIM = $93.56\%$ and PSNR = $20.89$ dB, effectively suppressing limited-view artifacts. \\
    \cmidrule{2-4}
    
    & NA Mechanism \cite{77chan2024local} 
    & Transformer-based sparse PAT reconstruction with neighborhood attention (NA) mechanism and sliding window spatial attention for enhanced local context modeling.
    & Achieves PSNR = $29.39$ dB and SSIM = $0.853$ at 16 projections, and average PSNR = $31.10$ dB / SSIM = $0.878$ across 16--64 projections, consistently outperforming UNet, NAFNet, Restormer, and Shuffleformer in extreme sparse PAT reconstruction. \\

    \bottomrule
    \end{tabularx}
    \end{table}
    \FloatBarrier

\begin{figure}[]
    \centering
    \includegraphics[width=13cm]{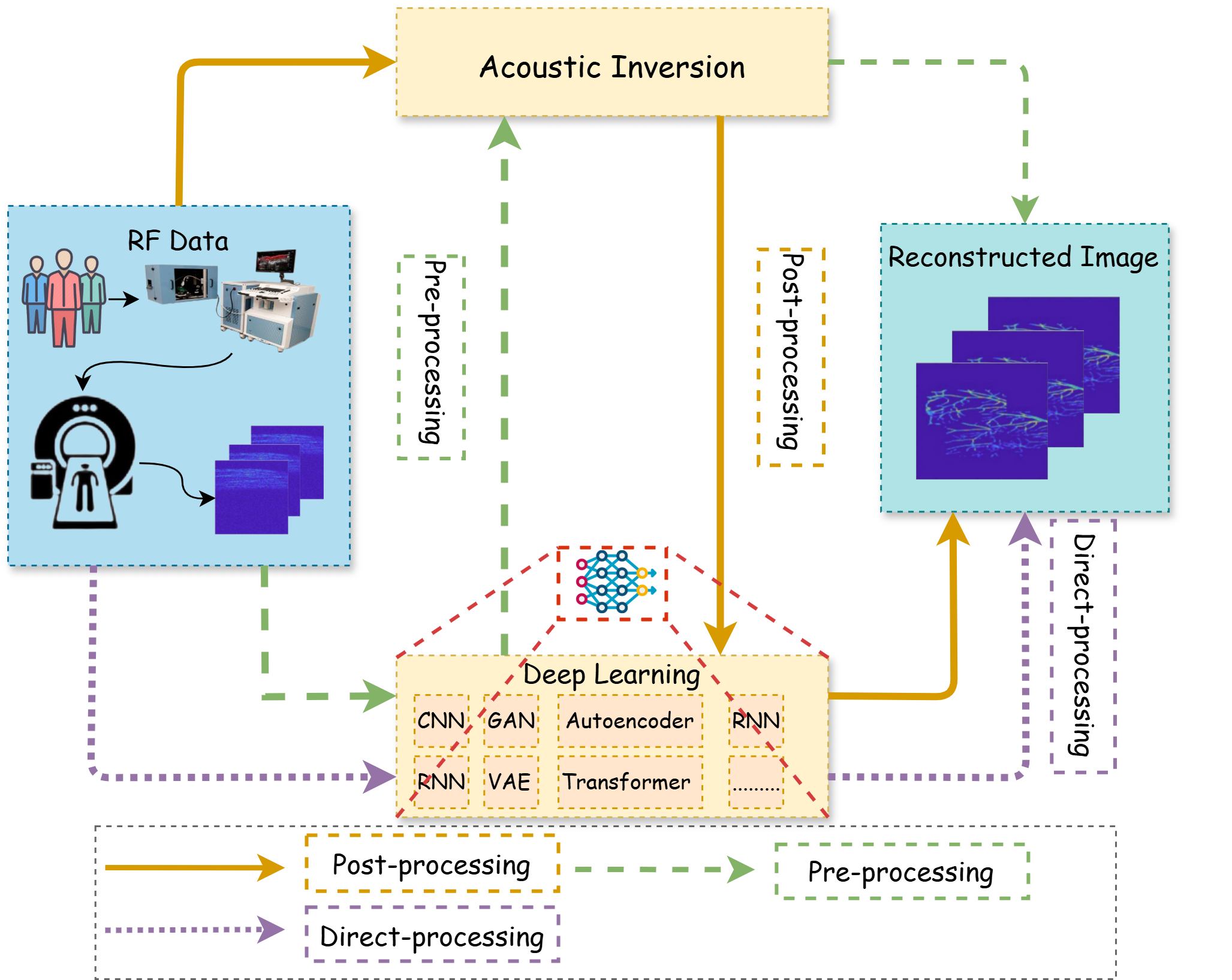}
    \caption{Paradigm illustration using DL for PACT reconstruction.
}
    \label{fig6}
\end{figure}

\subsubsection{DL-based Reconstruction During Pre-processing Stage}
Data preprocessing employs neural networks to optimize raw data for reconstruction, addressing limitations such as insufficient bandwidth and limited angle data. Specific networks process photoacoustic sinograms, thereby enhancing signal bandwidth and resolution. Gutta \textit{et al.} \cite{66gutta2017deep} utilized a fully connected deep neural network (FC-DNN) to correct sonograms, broaden the effective bandwidth, and increase the SNR of beamformed images by approximately 6 dB. Similar approaches were reported by other researchers  \cite{67awasthi2020deep} and \cite{68durairaj2020unsupervised}.

\subsubsection{DL-based Reconstruction Within Direct-processing Stage}
Direct processing in neural networks for PAI involves the use of architectures that map incoming photoacoustic signals directly to the resulting images, bypassing traditional step-by-step processing. Feng \textit{et al.} \cite{69feng2020end} introduced an end-to-end Res-Unet model that combined ResNet and Unet architectures to generate high-quality PA images from raw sensor inputs. The model features skip connections and a residual learning mechanism that enhances training efficiency and prevents network degradation. This approach produces images with clear edges and fewer artifacts, simplifying the imaging pipeline and potentially reducing computational requirements. Comparisons with traditional methods and other DL models can provide further insights into their performance and practical implications. This advancement holds promise for enhancing PAI in biomedical applications. 

\subsubsection{DL-based Reconstruction Throughout Post-processing Stage}
Postprocessing techniques enhance images from conventional algorithms, particularly by addressing the challenges of sparse sampling and limited-angle acquisition in photoacoustic imaging. Antholzer \textit{et al.} \cite{59antholzer2018photoacoustic} used a U-Net network to refine FBP-reconstructed images under limited-angle conditions. Guan \textit{et al.} \cite{70guan2019fully} developed a Fully Dense U-Net (FD-UNet) with dense connections to improve feature transmission. Zhang \textit{et al.} \cite{71zhang2021limited} introduced a dual-domain Unet (DuDoUnet) with an Information Sharing Block to reduce artifacts. Lan \textit{et al.}'s JEFF-Net \cite{72lan2023jointed} employs a DL fusion framework to suppress artifacts, whereas unsupervised methods, such as PA-GAN \cite{73zhang2019progressive} and PAT-AND \cite{74zhong2024unsupervised} address data labeling issues, with PAT-AND performing well under sparse angular conditions.

Physics-informed DL approaches integrate physical models with neural networks. Song \textit{et al.} \cite{75song2024accelerated} proposed a fast sparse reconstruction strategy that combines a multichannel denoising autoencoder prior to a model-based iteration. Guo \textit{et al.} \cite{76guo2022net} introduced a high-quality PACT strategy under limited-angle conditions using a fractional diffusion model to enhance imaging quality and speed.

Sparse sampling is crucial for rapid data acquisition in PAI; however, reconstructing high-quality images from sparse data is challenging. Chan \textit{et al.} \cite{77chan2024local} proposed a neighborhood attention mechanism for sparse PACT reconstruction that emphasized local neighborhood information. Tong \textit{et al.} \cite{78tong2020domain} developed a feature-projection network (FPNet) for direct reconstruction from sparse and limited-angle data; however, this requires significant computational resources. 

\subsubsection{DL-based Reconstruction in Hybrid-processing Stage}
Lan \textit{et al.}'s Y-Net was a hybrid DL framework that combines direct reconstruction with postprocessing to extract information from raw data and preliminary reconstructed images \cite{79lan2020net}. Inspired by the U-Net architecture, Y-Net reconstructs the initial PA pressure distribution by optimizing both the raw data and beamformed images using two encoder paths and one decoder path.

\subsection{PAM Image Reconstruction}
In PACT and PAM, DL is applied in different ways. In PAM, without requiring inverse reconstruction, DL models directly map input signals to output images, thereby enhancing image quality \cite{3wang2008tutorial}, \cite{59antholzer2018photoacoustic}. This approach effectively addresses the challenges in PAM, such as improving reconstruction, removing artifacts, denoising data, enhancing spatial resolution, and upsampling sparse data. In addition, DL efficiently approximates nonlinear spatial mapping using GPU acceleration. Existing reconstruction algorithms have been continuously refined. This section examines the impact of current DL models on photoacoustic image performance in PAM, focusing on resolution, noise reduction, and imaging depth. 

\subsubsection{DL for Resolution Enhancement in PAM Reconstruction}
Spatial resolution is crucial for evaluating photoacoustic images and is influenced by both hardware and reconstruction algorithms. Reducing spatial sampling density and increasing scanning step size in PAM to accelerate imaging degrades resolution and causes aliasing. DL has been applied to lower laser pulse energy and enhance undersampling in PAM.

Recently, a non-pretrained Deep Iterative Prior (DIP) model was developed to enhance undersampled PAM images and speed up imaging \cite{80vu2021deep}. This model iteratively optimizes fully sampled images from undersampled data using a known downsampling mask, thereby enhancing generalization without the need for paired training. It outperformed interpolation methods and pretrained DL models on mouse brain vasculature and bioprinted gel images without prior training data, marking its first application in PAM.

Seong \textit{et al.} \cite{81seong2023three} developed a DL-based method for 3D volumetric reconstruction in PAM, which reduced imaging time and data volume by fully reconstructing undersampled datasets. They adapted SRResNet and validated its performance, achieving the first DL-based 3D reconstruction of undersampled PAM data.

Li \textit{et al}. \cite{82li2024removing} introduced PSAD-UNet, a model that enhances transcranial PAI by mitigating bone plate effects, with broad applications in both preclinical and clinical settings. Wang \textit{et al.} \cite{83wang2024reconstructing} developed PADA U-Net, which reconstructs full images from undersampled data, overcoming the tradeoff between imaging speed and spatial resolution. Shahid \textit{et al.} \cite{84shahid2021batch} proposed BRn-ResNet, which improved the training stability of U-Net and reduced artifacts from sparse data through residual blocks, thereby achieving an SSIM of 0.97.

Cao \textit{et al.} \cite{85cao2025mean} proposed a mean-reverting diffusion model for AR-PAM by employing an iterative reverse-time SDE to balance imaging depth and lateral resolution, thereby enhancing imaging quality and broadening applicability without sacrificing depth. Le \textit{et al.} \cite{86le2023enhanced} implemented a computational strategy with two GANs for semi-/unsupervised high-resolution sensitive reconstruction in AR-PAM, maintaining imaging capability at enhanced depths. Cheng \textit{et al.} \cite{87cheng2022high} utilized DL for image transformation to improve deep penetration in OR-PAM.

Zhao \textit{et al.} \cite{88zhao2021deep} proposed a multitask residual dense network (MT-RDN) to enhance image quality in traditional OR-PAM at ultralow laser doses, incorporating multisupervision learning, dual-channel acquisition, and densely connected layers for richer reconstruction details. 

\FloatBarrier
\begin{table}[htbp]
\centering
\caption{Brief summary of DL approaches for PAM reconstruction.}
\label{table7}
\renewcommand{\arraystretch}{1.2} 
\setlength{\tabcolsep}{5pt}        
\footnotesize

\renewcommand{\tabularxcolumn}[1]{p{#1}}

\begin{tabularx}{\textwidth}{
    >{\raggedright\arraybackslash}p{2.6cm}
    >{\raggedright\arraybackslash}p{2.0cm}
    >{\raggedright\arraybackslash}X
    >{\raggedright\arraybackslash}X
}
    \toprule
    \textbf{Aims} & \textbf{Models} & \textbf{Backbones} & \textbf{Key Achievements} \\
    \midrule

    DL for Sparse Sampling Recovery
    & DIP~\cite{80vu2021deep} 
    & Untrained CNN leveraging deep image prior; iteratively optimizes a full image to match undersampled observation via known downsampling mask.
    & Recovers high-quality PAM images from only 1.4\% of fully sampled pixels; outperforms interpolation and rivals supervised DL methods without requiring training or ground truth. \\
    \cmidrule{2-4}

    & SRResNet~\cite{81seong2023three} 
    & Modified SRResNet with experimentally calibrated subpixel upsampling for 3D isotropic reconstruction from anisotropic sparse PAM volumes.
    & Achieves 80$\times$ faster imaging and 800$\times$ data reduction; outperforms interpolation across PSNR, SSIM, MSE, and 3D vascular fidelity metrics (VOP-MAP, VOP-3D) on real undersampled PAM data. \\

    \midrule

    \multirow[t]{4}{2.6cm}[0pt]{DL-Driven Resolution Enhancement in PAM Reconstruction}
    & PADA U-Net~\cite{83wang2024reconstructing} 
    & Photoacoustic Dense Attention U-Net (PADA U-Net) with dense connections and attention gates for enhanced feature recovery in undersampled OR-PAM.
    & At 4$\times$ undersampling, improves PSNR by +2.33 dB and SSIM by +0.117 over bilinear interpolation on bovine bone; enables high-resolution OR-PAM at high speed for bone microstructure imaging. \\
    \cmidrule{2-4}

    & BRn-ResNet~\cite{84shahid2021batch} 
    & ResNet enhanced with batch renormalization (BRn) to stabilize training and reduce subsampling artifacts in sparse-view PAT.
    & Delivers high-resolution PAT reconstructions with SSIM $\approx$ 0.97, even when trained on low-quality data; significantly improves training stability and artifact suppression under sparse sampling. \\
    \cmidrule{2-4}

    & Mean-reverting Diffusion Model~\cite{85cao2025mean} 
    & Mean-reverting diffusion model that learns the degradation prior from optical- to acoustic-resolution PAM and recovers high-resolution images via reverse sampling.
    & Achieves more than $3.6\times$ lateral resolution enhancement; improves PSNR by 66\% and SSIM by 480\% on in vivo AR-PAM data, enabling optical-level resolution with deep penetration. \\
    \cmidrule{2-4}

    & MT-RDN~\cite{88zhao2021deep} 
    & Multitask Residual Dense Network (MT-RDN) with multisupervised learning, dual-channel input, and balanced task weighting for joint denoising, super-resolution, and vascular enhancement.
    & Achieves high-quality OR-PAM at 32$\times$ lower laser dosage (within ANSI limit) and 4$\times$ undersampling; enables real-time, cross-system, cross-anatomy imaging with superior denoising and resolution---first method to jointly resolve the laser dose--speed--quality trilemma. \\

    \midrule

    \multirow[t]{2}{2.6cm}[0pt]{DL for Image Denoising in PAM Reconstruction}
    & UPAMNet~\cite{90liu2024upamnet} 
    & UPAMNet: Unified attention-enhanced CNN with pixel- and perception-level mixed loss for joint PAM super-resolution and denoising.
    & Improves PSNR by +0.59 dB (4$\times$ undersampling), +1.37 dB (16$\times$ undersampling), and +3.9 dB (denoising); generalizes well across multiple PAM datasets with a unified reconstruction framework. \\
    \cmidrule{2-4}

    & PnP Prior~\cite{93zhang2023adaptive} 
    & Plug-and-Play (PnP) framework combining a deep CNN prior with model-based optimization using depth- and frequency-dependent PSF kernels for adaptive AR-PAM enhancement.
    & Outperforms all methods in simulations (best PSNR/SSIM); boosts in vivo SNR from 6.34 to 35.37 and CNR from 5.79 to 29.66; enables a single model to handle diverse AR-PAM degradation conditions via physical PSF integration. \\
    \midrule

    \multirow[t]{2}{2.6cm}[0pt]{DL for Imaging Depth Enhancement in PAM Reconstruction}
    & ResUnet-AG~\cite{94meng2022depth} 
    & Two-stage residual U-Net with attention gates for adaptive out-of-focus AR-PAM image restoration.
    & Extends AR-PAM depth-of-focus from 1 to 3 mm; recovers near-diffraction-limited resolution at 2 mm off-focus, enabling high-quality deep-tissue imaging with standard AR-PAM systems. \\
    \bottomrule
\end{tabularx}
\end{table}
\FloatBarrier

\renewcommand{\tabularxcolumn}[1]{m{#1}}
\begin{table}[htbp]
\centering
\caption{Brief summary of DL-based reconstruction approaches for PAE.}
\label{table8}
\renewcommand{\arraystretch}{1.5}
\setlength{\tabcolsep}{6pt}
\footnotesize
\begin{tabularx}{\textwidth}{%
    >{\raggedright}m{3.2cm}
    >{\raggedright}m{4.8cm}
    >{\raggedright\arraybackslash}X
}
\toprule
\textbf{Models} & \textbf{Backbones} & \textbf{Key Achievements} \\
\midrule
Focus U-Net \cite{97yeung2021focus} 
& This study incorporated several architectural modifications into the Focus U-Net, including the addition of short-range skip connections and deep supervision. 
& Achieves SOTA DSC of 0.941 (CVC-ClinicDB), 0.910 (Kvasir-SEG), and 0.878 (5-dataset combined); 14–15\% improvement over previous SOTA, enabling reliable computer-aided polyp detection. \\
\midrule
Deep EMI Denoising \cite{98gulenko2022deep} 
& This study proposed a modified U-Net for EMI denoising in OR-PAE, trained on in vivo rat and rabbit data acquired with 40 MHz transducers.
& Enables clear 3D reconstruction of \textasciitilde50\,\textmu m capillary networks by removing EMI noise; outperforms classical filters and other CNNs; offers a scalable AI solution for low-SNR, low-cost PAE/PAT systems. \\
\midrule
Contrast Limited Adaptive Histogram Equalization (CLAHE) \cite{99moghtaderi2024endoscopic} 
& This study proposed a method to enhance endoscopic images by generating three complementary sub-images using Contrast Limited Adaptive Histogram Equalization (CLAHE), image brightening, and detail enhancement. 
& Significantly improves global and local contrast while suppressing noise amplification, enabling clearer visualization of mucosal and vascular structures in PAE. \\
\midrule
Approximate Gaussian acoustic field \cite{100wang2022photoacoustic} 
& This study proposed a new photoacoustic/ultrasound endoscopic imaging reconstruction algorithm based on the approximate Gaussian acoustic field, which significantly improves the resolution and SNR of the out-of-focus region. 
& Achieves up to 92.6\% (simulation) and 52.3\% (phantom) improvement in lateral resolution, SNR boost from 16\,dB to 36\,dB, 8\,s/frame reconstruction speed, and validated superior image quality in in vivo rabbit rectal endoscopy. \\
\midrule
Improved back-projection (IBP) algorithm \cite{101xiao2022acoustic} 
& This study developed an IBP algorithm for focused detection over centimeter-scale imaging depth, alongside a deep-learning-based algorithm to remove electrical noise. 
& Achieves 0.77/0.65\,mm lateral resolution and 25/38\,dB SNR at 1.4\,cm depth; enables molecular discrimination of ICG in the rabbit rectum; DL denoising eliminates the need for averaging; identifies 800\,nm as the optimal wavelength for deep-tissue imaging. \\
\bottomrule
\end{tabularx}
\end{table}

\subsubsection{DL for Image Denoising in PAM Reconstruction}
Noise significantly affects photoacoustic image reconstruction, causing artifacts due to hardware limitations and phase mismatches resulting from inhomogeneous sound velocity and tissue scattering. Zhou \textit{et al.} \cite{89zhou2019noise} introduced a denoising method that combines empirical mode decomposition (EMD) and conditional mutual information, where EMD decomposes noisy signals into intrinsic mode functions (IMFs), and conditional mutual information aids in denoising.

Liu \textit{et al.}~\cite{90liu2024upamnet} proposed UPAMNet, a deep-learning framework that leverages deep image priors to simultaneously address super-resolution and denoising in PAM. By integrating attention-enhanced feature modules and a hybrid training objective combining pixel-wise and perceptual losses, the method achieved notable improvements of 0.59\,dB and 1.37\,dB PSNR gains for 1/4 and 1/16 undersampled data, respectively, along with a 3.9\,dB SNR boost during denoising. More importantly, the model demonstrated strong generalization via few-shot and zero-shot transfers to unseen datasets, indicating its potential for practical clinical deployment. The overall architecture and representative results are shown in Fig. ~\ref{fig7}.


In photoacoustic imaging, the spatial resolution is degraded by factors such as ultrasound attenuation, phase deviations from sound speed variations, and signal waveform broadening. He \textit{et al.} \cite{91he2022noising} addressed this issue by proposing an attention-enhanced GAN with an improved U-net generator to remove noise from PAM images. 

However, the low spatial resolution of AR-PAM imaging has limited its adoption \cite{92zhang2023organ}. Previous models lacked flexibility or required complex priors and could not adapt to different degradation models. Zhang \textit{et al.} \cite{93zhang2023adaptive} developed a deep CNN-based model that adaptively handles various degradation functions in AR-PAM image enhancement. This model improves PSNR, SSIM, SNR, and CNR in both simulated and in vivo images by learning vascular image statistics as a plug-and-play prior, demonstrating its interpretability, flexibility, and broad applicability.

\begin{figure}[]
    \centering
    \includegraphics[width=13cm]{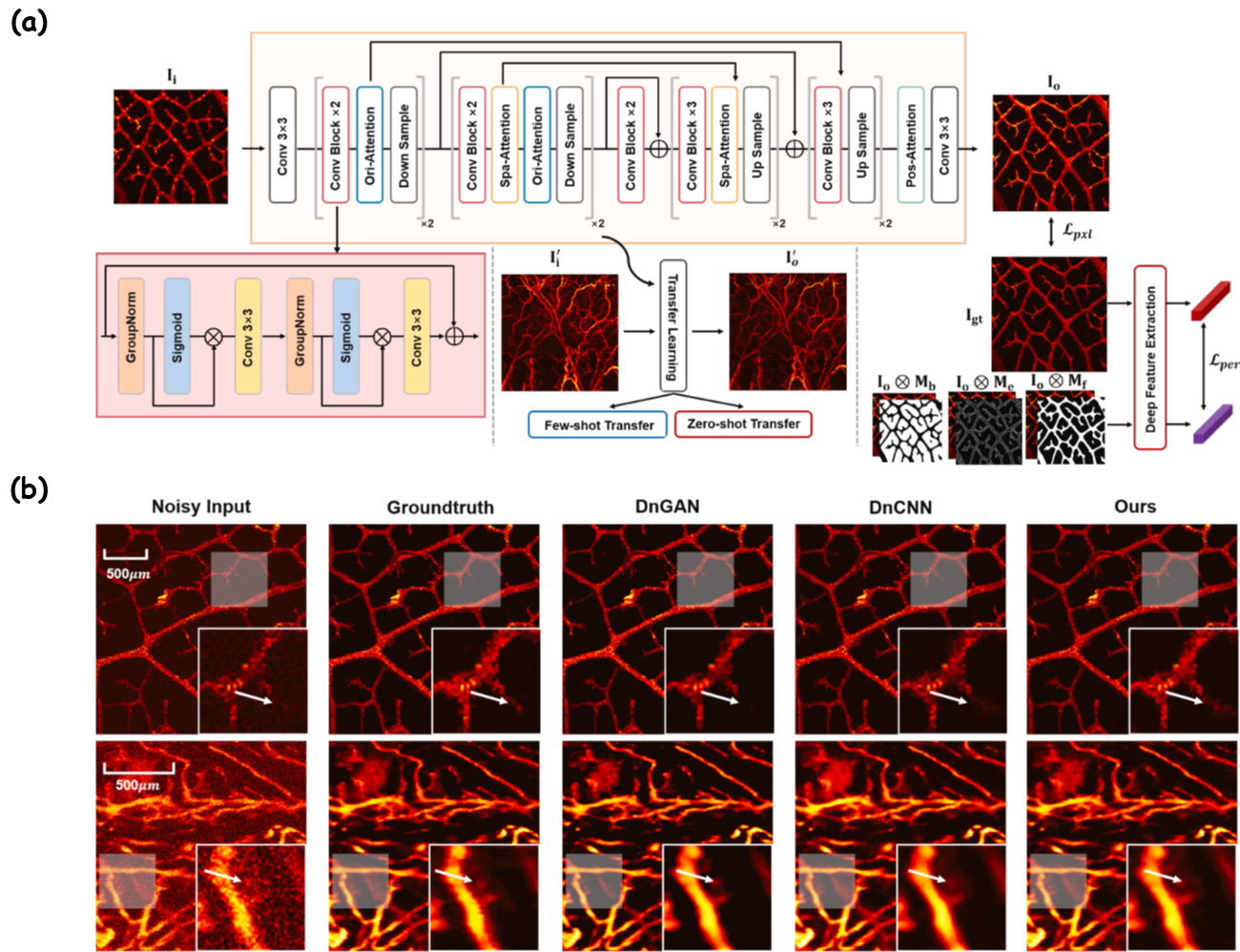}
    \caption{(a) Illustration of UPAMNet architecture, adapted from~\cite{90liu2024upamnet}. The network comprises three modules: feature contraction, feature connection, and feature expansion. Three attention blocks, designed based on deep image priors, are integrated to enhance feature learning. A combined training constraint leveraging semantic segmentation is applied at both pixel and perceptual levels. The inset shows the detailed structure of the ResConv block. Few-shot and zero-shot transfer learning are employed to evaluate generalization on unseen datasets; (b) Denoising results via supervised learning adapted from~\cite{90liu2024upamnet}: first row---Dataset D-I; second row---Dataset D-II. Zoomed views of the white shaded regions (lower right) are shown.}
    \label{fig7}
\end{figure}

\subsubsection{DL for Imaging Depth Enhancement in PAM Reconstruction}
In clinical settings, imaging of deep tissues is essential, and neural networks play a key role in enhancing imaging depth.

Traditional AR-PAM systems face challenges such as poor reconstruction in deep or out-of-focus regions. Meng \textit{et al.} developed ResUnet-AG \cite{94meng2022depth} a Residual U-Net with an attention mechanism that enhances deeper tissue imaging by ignoring background noise and extending the depth of field from 1 to 3 mm. Table \ref{table7} summarizes the PAM reconstruction approaches.

\subsection{ PAE Image Reconstruction }
Endoscopy is essential for diagnosing internal organ diseases; however, it is often difficult to differentiate microvessels or identify lesions because of internal environmental and imaging conditions \cite{13guo2020photoacoustic}. PAE faces challenges owing to limited fields of view and complex in vivo environments. Factors such as blood, illumination variations, specular reflections, and smoke introduce noise, degrading the image quality, particularly in occluded regions. Enhancing the endoscopic image quality by improving details, contrast, brightness, and removing specular reflections is crucial \cite{96li2024endosrr}.

Yeung \textit{et al.} introduced Focus U-Net \cite{97yeung2021focus}, which is a dual attention-gated deep neural network that selectively learns polyp features by incorporating skip connections, deep supervision, and Hybrid Focal Loss to address class imbalance. This network holds promise for noninvasive colorectal cancer screening and other biomedical image segmentation tasks.

In clinical PAE, electromagnetic interference (EMI) noise severely degrades the image quality, particularly in low-cost or miniaturized systems with limited SNR. Gulenko \textit{et al.}~\cite{98gulenko2022deep} proposed a deep-learning-based algorithm to address this challenge by leveraging a modified U-Net architecture trained on in vivo photoacoustic data from the rat colorectum and rabbit urethra acquired with 40\,MHz transducers. Their approach effectively removed structured EMI artifacts while preserving fine vascular structures, outperforming classical filtering methods and alternative CNNs such as SegNet and FCN variants. This method enables high-fidelity 3D visualization of microvasculature, including mesh-like capillary networks (\SI{\sim 50}{\micro\meter}) and offers a scalable software solution for EMI suppression in emerging LED- or laser-diode-based photoacoustic tomography systems, where hardware-based noise mitigation is impractical.

Building on this, Moghtaderi \textit{et al.} \cite{99moghtaderi2024endoscopic} developed a technique that generates three complementary subimages using contrast-limited adaptive histogram equalization (CLAHE), image brightening, and detail enhancement. These subimages are decomposed via multilevel wavelet transforms and guided filters and then fused with weights to produce a final enhanced image suitable for low-light conditions in endoscopy.

Improving the resolution and SNR in the defocused regions is crucial for designing advanced ultrasound/PAE systems. Wang \textit{et al.} \cite{100wang2022photoacoustic} introduced a novel reconstruction algorithm based on an approximate Gaussian acoustic field that enhanced the resolution and SNR in out-of-focus areas. This algorithm was validated using a chicken breast phantom, and in a rabbit rectal endoscopy experiment, it demonstrated improved imaging quality, faster dynamic focusing, and enhanced overall imaging performance.

Molecular PAE imaging of deep tissues presents significant technical challenges. Xiao \textit{et al.} \cite{101xiao2022acoustic} developed an IBP algorithm for focused detection over centimeter-scale imaging depths and employed a DL-based method to reduce electrical noise from the step motor. These advances have reduced the scanning time and fostered the development of high-penetration molecular PAE. Table \ref{table8} summarizes PAE reconstruction approaches.

\subsection{Reliability and Generalization Challenges of DL in PAI}

Despite their remarkable performance in controlled settings, DL-based methods in PAI face several critical limitations that hinder robust clinical deployment. A primary concern is overfitting; models trained on limited or non-representative datasets often fail to generalize across different imaging systems, patient populations, or acquisition protocols, leading to degraded performance in real-world scenarios \cite{168antun2020instabilities}. Moreover, DL approaches are inherently data hungry, requiring large-scale, high-quality, and expertly annotated datasets---resources that remain scarce in photoacoustic imaging, particularly for rare pathologies or multimodal fusion tasks \cite{169yang2023deep}. The “black-box” nature of most DL architectures further complicates their clinical adoption because a lack of interpretability undermines trust and hampers regulatory approval in safety-critical applications \cite{170montavon2019explainable}. Most critically, when trained on biased or incomplete data, DL models may generate physically implausible reconstructions---such as spurious structures in low-signal regions---that mimic the true pathology but lack biological plausibility \cite{171hauptmann2020deep}. To ensure reliable use, users must validate models on truly independent test sets, monitor performance across diverse data distributions, and employ uncertainty quantification techniques (e.g., Monte Carlo dropout) to identify low-confidence predictions \cite{172kendall2017uncertainties}. Crucially, DL outputs should be cross-validated against physics-based reconstructions or expert interpretations, particularly when clinical decisions depend on the results. Thus, while DL offers transformative potential for accelerating and enhancing photoacoustic image reconstruction, its clinical integration demands rigorous validation, transparency, and a clear understanding of its inherent limitations. The clinical adoption of DL-enhanced PAI also depends on practical factors such as laser safety, acoustic coupling stability, imaging speed, and system cost. Although these operational barriers are well-documented in the literature \cite{18beard2011biomedical}, \cite{152wang2012photoacoustic}, they underscore that even a perfectly robust DL model cannot succeed in isolation from a broader imaging ecosystem.

\section{DL-based Quantitative Analysis in Photoacoustic Imaging}
Although PAI excels in generating high-contrast images, its translation from a qualitative technique to a reliable clinical diagnostic tool hinges on accurate quantitative analysis. The accurate quantification of physiological parameters---such as blood oxygen saturation (sO$_2$)---is critical for clinical decision-making in oncology, vascular diseases, and neurology.

PAI provides comprehensive structural, functional, molecular, and kinetic information by leveraging endogenous chromophores---such as hemoglobin, lipids, melanin, and water---as well as exogenous contrast agents, including clinically approved dyes such as ICG and methylene blue. A key feature of PAI is its ability to differentiate deoxyhemoglobin (Hb) from oxyhemoglobin (HbO$_2$), enabling noninvasive mapping of blood oxygen saturation (sO$_2$)---crucial for detecting conditions such as ischemia, hypoxia, and hypoxemia \cite{102yang2019quantitative}. Furthermore, PAI utilizes the strong optical absorption of melanin to visualize melanomas and other pigmented lesions, offering valuable diagnostic insights \cite{20wang2009multiscale}.

However, accurate interpretation of this rich contrast is hindered by the spatially heterogeneous optical fluence in biological tissues and significant spectral overlap among chromophores. Consequently, robust quantitative analysis is essential and typically requires optical fluence compensation and spectral unmixing to recover accurate chromophore concentrations \cite{103zhang2024navigating}. This section focuses on the pivotal role of DL in addressing these fundamental challenges, as summarized in Fig.~\ref{fig8}, enabling robust and clinically relevant quantitative photoacoustic imaging (qPAI).

\begin{figure}[]
    \centering
    \includegraphics[width=13cm]{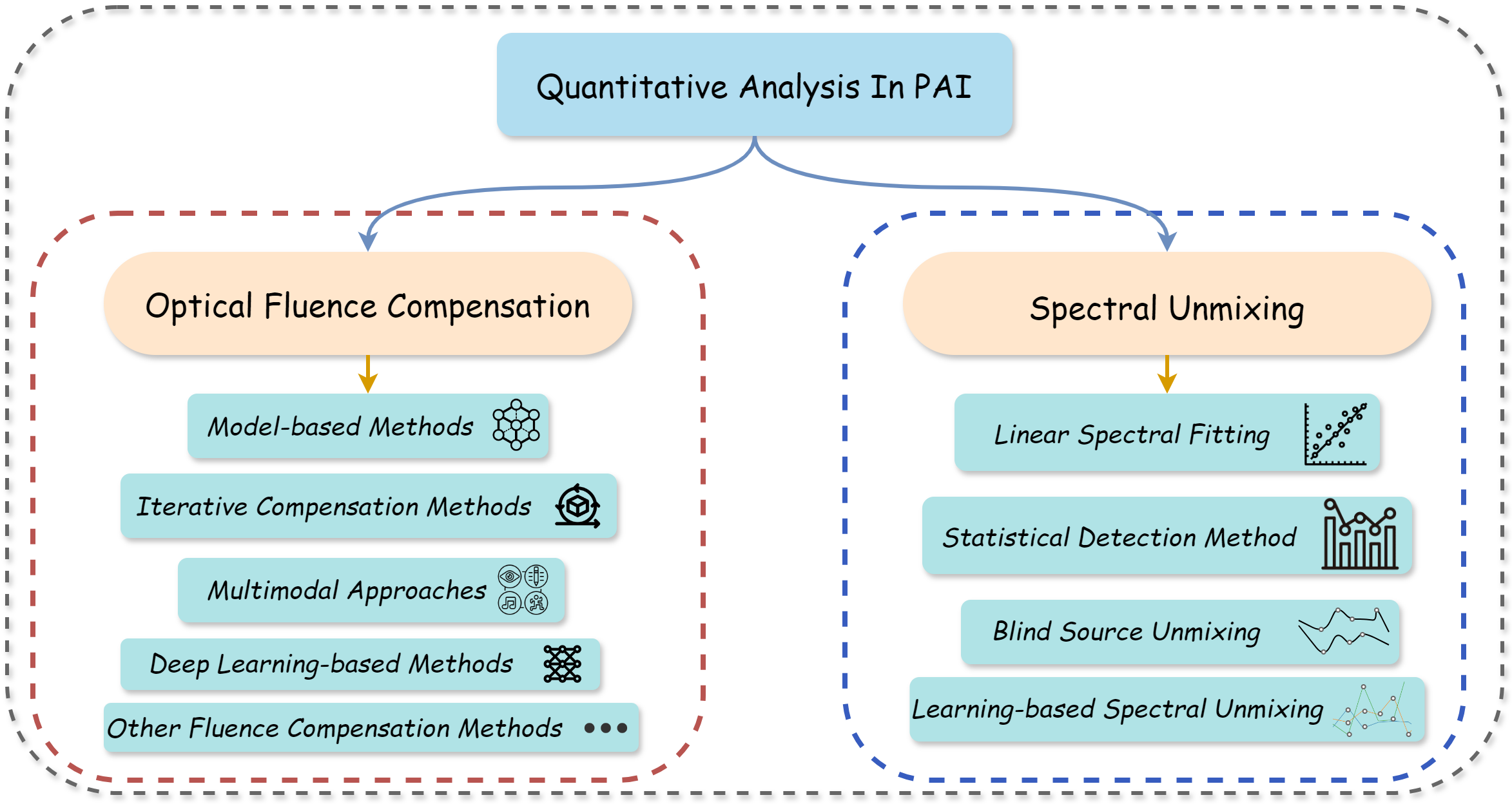}
    \caption{Overview of quantitative analysis of PAI.}
    \label{fig8}
\end{figure}

\subsection{Optical Fluence Compensation}
The accuracy and clinical utility of qPAI are fundamentally limited by the spatial distribution of optical fluence, which varies with tissue depth and illumination wavelength. Optical fluence compensation mitigates quantitative inaccuracies arising from inhomogeneous light distribution, thereby enhancing the accuracy of estimating physiological parameters, such as hemoglobin concentration and blood oxygen saturation (sO$_2$). DL models can be trained to learn complex feature representations from raw photoacoustic data and directly estimate the underlying optical fluence, thereby enabling data-driven compensation for variations in tissue optical properties. This data-driven approach enables automatic and efficient fluence compensation, thereby significantly enhancing the quantitative accuracy of reconstructed PA images. Although various fluence compensation strategies exist---including model-based, iterative, and multimodal approaches---DL–based methods have recently emerged as powerful data-driven alternatives \cite{103zhang2024navigating}.

\subsubsection{Model-based Methods}
Several light transport models have been employed to estimate the optical fluence, including the diffusion equation (DE) and Monte Carlo (MC) simulations, which numerically solve the radiative transfer equation (RTE) \cite{104dantuma2021tunable} and diffusion dipole model (DDM) \cite{105zhou2020evaluation}. The diffusion equation is well-suited for macroscopic fluence estimation in highly scattering tissues, as demonstrated by Zhao \textit{et al.}, who utilized finite element analysis (FEA) to generate fluence maps \cite{106zhao2017optical}. Monte Carlo simulations employed by Kirillin \textit{et al.} \cite{107kirillin2017fluence} and Hirasawa \textit{et al.} \cite{108hirasawa2016multispectral} stochastically modeled photon propagation and effectively mitigated spectral coloring artifacts caused by wavelength-dependent light attenuation. Ranasinghesagara \textit{et al.} \cite{109ranasinghesagara2014reflection} employed multi-illumination photoacoustic microscopy (MI-PAM) to estimate the tissue optical properties, which were subsequently used as inputs to an MC model for fluence calculation.

\subsubsection{Iterative Compensation Methods}
Iterative fluence compensation estimates the optical fluence distribution by solving an inverse problem; starting from an initial guess, it iteratively updates the fluence to minimize the discrepancy between the measured and simulated photoacoustic signals until convergence. Unlike model-based approaches, iterative methods do not require prior knowledge of the tissue optical properties, thereby avoiding potential biases introduced by inaccurate parameter assumptions. Consequently, they often exhibit greater robustness in heterogeneous or previously uncharacterized tissue environments. These algorithms are well suited for complex, heterogeneous media but are computationally intensive---particularly for 3D datasets or when coupled with Monte Carlo (MC) light transport models \cite{110jin2019single}. While reducing the computational cost is desirable, achieving higher quantitative accuracy often requires additional iterations or more sophisticated optimization schemes, highlighting the inherent trade-off between accuracy and computational efficiency.

\subsubsection{Multimodal Approaches}}

Multimodal integration is pivotal for the clinical translation of PAI, as it synergistically combines complementary modalities to overcome the limitations of any single technique, which cannot simultaneously provide high-resolution anatomical detail, deep tissue penetration, functional specificity, and molecular sensitivity. The most established configuration integrates PAI with US, enabling co-registered anatomical (US) and functional/molecular (PAI) imaging using a shared transducer array. This design underpins the commercial systems for breast and thyroid imaging. Recent advancements in lightweight handheld scanners have enabled clinical panoramic volumetric PA/US imaging, facilitating 3D visualization of vascular anatomy, hemodynamics, and oxygen saturation over large fields of view \cite{165lee2023panoramic}.

Beyond US, hybrid platforms incorporating OCT deliver micrometer-scale structural information in superficial tissues (e.g., skin, retina), complementing PAI's hemodynamic contrast for dermatological and ophthalmic applications. For deep-tissue imaging, integration with MRI leverages superior soft-tissue contrast to guide PAI reconstruction and interpretation, particularly in neuro-oncology. Emerging systems incorporating fluorescence lifetime imaging (FLIM) enable multiparametric mapping of absorption- and fluorescence-based biomarkers in a single acquisition.

\begin{figure}[htbp]
\centering
\includegraphics[width=\linewidth]{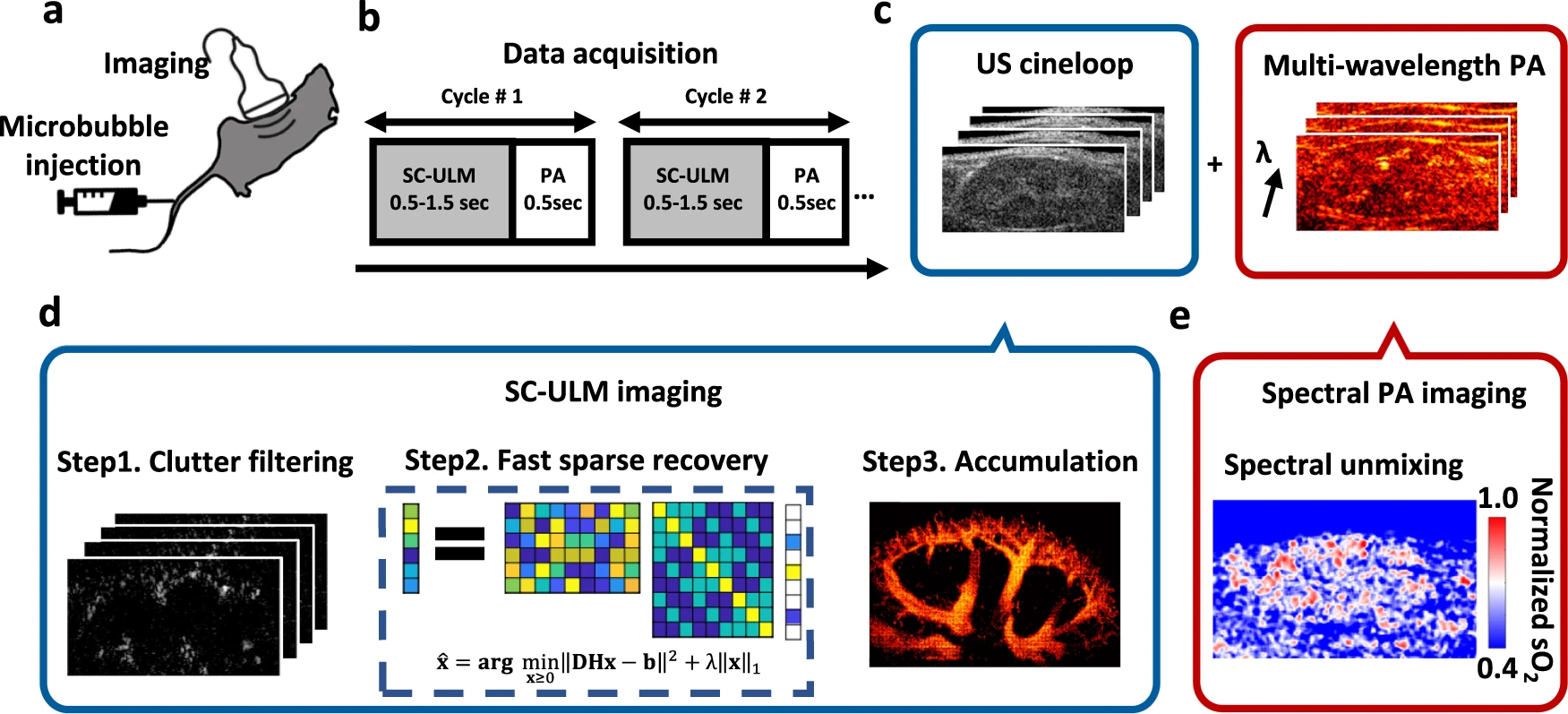}
\caption{Representative multimodal imaging workflow combining photoacoustic and super-resolution ultrasound, adapted from Zhao \textit{et al.}~\cite{181zhao2023hybrid}. 
(a) Animals are injected with a microbubble solution before (or during) imaging. 
(b) During data acquisition (DAQ), multi-wavelength photoacoustic (PA) and plane-wave fast ultrasound imaging (US) are recorded alternately at each position; the minimum DAQ duration for ultrasound localization microscopy (ULM) is determined by the pixel saturation curve characterization time. 
(c) Ultrasound (US) cine-loops and multi-wavelength PA images are stored and processed separately via the pipelines shown in (d) and (e). 
(d) Sparsity-constrained (SC)-ULM imaging involves three steps: clutter filtering, sparse recovery over frames, and location accumulation. 
(e) Blood oxygen saturation (sO$_2$) maps are generated through linear spectral unmixing of multi-wavelength PA images.}
\label{fig9}
\end{figure}

To concretely illustrate how such hybrid PA/US systems operate in practice, Fig.~\ref{fig9} shows a representative multimodal imaging workflow adapted from Zhao \textit{et al.}~\cite{181zhao2023hybrid}. This example demonstrates the synchronized acquisition and parallel processing of photoacoustic and super-resolution ultrasound data, enabling the simultaneous mapping of microvascular perfusion and blood oxygen saturation within a unified imaging framework.

From a quantitative perspective, auxiliary modalities enhance the accuracy of PAI-derived biomarkers. Ultrasound-derived structural priors constrain fluence models to improve the quantification of oxygen saturation (sO$_2$), whereas diffuse optical tomography (DOT) provides three-dimensional optical property maps for deep-tissue fluence correction. However, these approaches require additional hardware, labor-intensive segmentation (e.g., manual or AI-assisted delineation), and substantial computational resources---particularly for three-dimensional implementations. Despite these challenges, the synergistic value of multimodal systems in delivering comprehensive diagnostic information within clinical workflows underscores their critical role in the regulatory approval and clinical translation of PAI. In addition, calibration techniques such as reference phantom normalization and wavelength-dependent fluence correction are crucial to ensure accurate quantitative assessment.

\subsubsection{DL-based Methods}
DL offers a data-driven framework for optical fluence compensation in quantitative photoacoustic imaging. For instance, Yang \textit{et al.} \cite{102yang2019quantitative} employed a deep residual recurrent U-Net (DR2U-Net) to estimate sO$_2$ from photoacoustic data. However, the accuracy and generalizability of deep-learning models depend heavily on the availability of large and diverse training datasets. Moreover, training such models requires substantial computational resources and time, whereas complex architectures may incur high inference latency.

To mitigate these challenges---particularly data scarcity and limited generalizability---recent advances have integrated physical priors and leveraged multimodal data. Physics-informed DL frameworks now enable end-to-end quantitative reconstruction of absorption coefficients while considering the underlying photoacoustic physics \cite{166zheng2025physics}. Furthermore, models trained on co-registered ultrasound and photoacoustic data can estimate spatially varying fluence for improved sO$_2$ quantification \cite{167liang2024deep}. Self-supervised methods have demonstrated robust artifact reduction in clinical settings without requiring paired training data \cite{148lan2024masked}.

\subsubsection{Other Fluence Compensation Methods}
In addition to iterative and model-based approaches, alternative strategies, such as wavefront shaping and algorithmic correction, have been developed to address optical fluence variations. For instance, Caravaca-Aguirre \textit{et al.} demonstrated the use of a spatial light modulator (SLM) to modulate the wavefront of a 532\,nm laser, enabling focused light delivery through a scattering medium (a glass diffuser) onto capillary targets \cite{112caravaca2013high}. Complementarily, Fadhel \textit{et al.} proposed a fluence matching algorithm designed to correct errors in raw photoacoustic images caused by wavelength-dependent fluence variations, thereby improving the accuracy of sO$_2$ measurements \cite{113fadhel2020fluence}. These methods represent complementary strategies for mitigating fluence-related artifacts and enhancing the quantitative accuracy of PAI.

\subsection{Spectral Unmixing}
Spectral unmixing is a technique that decomposes a mixed photoacoustic (PA) signal within each pixel into contributions from individual chromophores---such as oxy- and deoxyhemoglobin, lipids, and melanin---by determining the spectral contribution of each endmember to the observed spectrum \cite{116shi2014incorporating}. PA data are typically acquired across multiple wavelengths, yielding multispectral images that encapsulate the overlapping absorption from these endmembers \cite{114dolet2021vitro}, \cite{115arabul2019unmixing}. Despite the complexity of tissue composition, spectral unmixing has proven effective in distinguishing major chromophores.

Applying DL to spectral unmixing can further enhance the separation of absorbers by learning complex nonlinear relationships in multispectral data, thereby enabling a more accurate quantitative analysis. Nevertheless, conventional spectral unmixing methods remain challenged by wavelength-dependent fluence variations, spectral cross-talk, and the lack of reliable calibration standards.

\subsubsection{Linear Spectral Fitting}
Traditional approaches to spectral unmixing in PAI typically rely on linear regression using the linear mixing model (LMM), which assumes that the measured signal at each pixel is a linear combination of the known absorption spectra of the constituent chromophores weighted by their respective concentrations \cite{117tzoumas2017spectral}. For example, Feng \textit{et al.} applied linear spectral unmixing to multi-wavelength photoacoustic (MWPA) data to isolate collagen as a biomarker for bone health assessments \cite{118feng2021detection}. Despite its widespread use, conventional linear spectral fitting is limited by the need for accurate prior knowledge of chromophore absorption spectra and wavelength-dependent fluence variations, which distort spectral signatures in complex or deep tissues \cite{115arabul2019unmixing}.

\subsubsection{Statistical Detection Methods}
Statistical detection methods aim to identify specific target chromophores with known absorption spectra---such as exogenous nanoparticles, reporter genes, and oxy- and deoxy-hemoglobin---without requiring a priori knowledge of the background absorption spectra. For example, Tzoumas \textit{et al.} \cite{119tzoumas2016eigenspectra} introduced eigenspectra multispectral optoacoustic tomography (eMSOT) to estimate sO$_2$ levels in vivo. Nevertheless, these approaches typically demand dense spectral sampling and high signal fidelity, which can increase computational complexity and pose practical challenges in dynamic in vivo imaging scenarios.

\subsubsection{Blind Source Unmixing}
Blind source unmixing (BSU) offers an alternative paradigm for qPAI that circumvents the need for prior knowledge of chromophore absorption spectra---a key limitation of conventional linear unmixing. Unlike model-based approaches, BSU aims to simultaneously estimate unknown spectral signatures and their spatial concentration maps directly from multispectral photoacoustic data, typically by exploiting statistical or structural properties, such as source independence, sparsity, or non-negativity \cite{120karoui2012blind}. This makes it particularly attractive for analyzing complex biological tissues, for which complete spectral priors are unavailable. However, its practical application in PAI remains challenging because of wavelength-dependent fluence variations, strong spectral correlations among endogenous chromophores (e.g., oxy- and deoxyhemoglobin), and the limited number of available excitation wavelengths.

\subsubsection{Learning-based Spectral Unmixing}
Kirchner \textit{et al.} developed context-enhanced quantitative photoacoustic imaging (CE-qPAI) using Monte Carlo simulations and random forests to enhance spectral unmixing \cite{121kirchner2018context}. Cai \textit{et al.} introduced an end-to-end ResUnet approach for accurate quantitative imaging using simulations. Gröhl’s team used learned spectral decoloring (LSD) with a neural network to improve oxygenation estimation accuracy \cite{122cai2018end}. Olefir \textit{et al.} \cite{123olefir2020deep} enhanced MSOT with DL-eMSOT, achieving superior performance using synthetic data.

To address the challenges of qPAI, it is essential to account for spatial variations in optical fluence to derive accurate absorption maps. Strategies include model-based methods, iterative compensation, multimodal approaches, and deep-learning techniques. Robust spectral unmixing is critical to obtaining accurate physiological and biochemical information at depth. Recent advancements such as linear spectral fitting, statistical detection methods, blind source unmixing, and learning-based approaches have significantly improved the decoding of mixed-pixel information into distinct chromophores. Notably, the challenges in qPAI are often interconnected owing to the hybrid nature of optical absorption and acoustic detection, which is beneficial for developing effective strategies to enhance the reliability and applicability of qPAI across various biomedical contexts.

\subsection{Emerging Excitation and Computational Strategies in PAI}
\label{subsec:emerging_strategies}

Recent progress in PAI stems from synergistic advances in light delivery, computational reconstruction, and wavefront control. These innovations address the fundamental trade-offs in PAI---between depth and resolution, speed and fidelity, cost, and performance---accelerating its translation to preclinical and clinical applications. Emerging strategies fall into two complementary categories: (i) novel excitation sources that enhance spectral coverage and practicality and (ii) advanced computational and light-field methods that improve image quality and effective optical focusing.

\subsubsection{Novel Excitation Sources}
\label{subsubsec:novel_sources}

Conventional PAI systems often rely on tunable optical parametric oscillators (OPOs) or Q-switched lasers, which deliver high pulse energy but are bulky, expensive, and limited in wavelength agility. Next-generation light sources overcome these constraints by enhancing tissue penetration, expanding the spectral range, and enabling compact system design.

Lasers operating in the second near-infrared window (NIR-II, 1000--1700~nm) leverage reduced tissue scattering and absorption within this band, enabling deep-tissue imaging with high spatial resolution, as demonstrated in breast tomography and transcranial vascular mapping~\cite{153yin2021organic}, \cite{154gao2025nir}. Combining NIR-II excitation with exogenous contrast agents, such as semiconducting polymers, further extends molecular imaging capabilities.

Supercontinuum sources, generated by pumping photonic crystal fibers with ultrashort pulses, provide broad emission spectra and enable rapid, all-electronic wavelength tuning, eliminating moving parts. These sources support real-time multispectral imaging to quantify key chromophores, including hemoglobin, lipids, and water~\cite{155dasa2019towards}, \cite{157dasa2020all}, \cite{156shaked2023label}.

High-power light-emitting diodes (LEDs) and laser diodes (LDs) offer compact, cost-effective alternatives that can be engineered for eye-safe operation. Advances in array architectures, beam shaping, and signal averaging have enabled portable PAI systems that are suitable for point-of-care diagnostics and wearable monitoring~\cite{158zhu2020towards}. Together, these sources broaden the applicability of PAI from deep tissue organ imaging to decentralized health monitoring.

\subsubsection{Advanced Computational and Light-Field Techniques}
\label{subsubsec:computational_lightfield}

Computational methods in PAI integrate physical models with data-driven priors to mitigate challenges such as limited-view detection, sparse sampling, and measurement noise. Model-based iterative reconstruction combined with DL---including unrolled networks and diffusion models---enables high-fidelity image recovery while reducing acquisition time and hardware complexity~\cite{159hauptmann2018model}, \cite{160deng2021deep}. Self-supervised and transfer-learning strategies further facilitate quantitative imaging in settings where ground truth annotations are unavailable.

Light-field modulation techniques, employing spatial light modulators (SLMs), digital micromirror devices (DMDs), or programmable metasurfaces, allow precise control over the optical wavefront. Structured illumination suppresses the out-of-focus background, enhancing contrast in scattering media~\cite{161murray2016super}. Meanwhile, wavefront shaping and time-reversed ultrasonically encoded (TRUE) focusing can confine the optical excitation to spots smaller than the acoustic diffraction limit, thus enabling high-resolution imaging deep within biological tissues ~\cite{162manohar2011gold}.

The integration of advanced light sources, adaptive illumination, and computational reconstruction represents a paradigm shift for PAI---transforming it from a passive imaging modality into an active, closed-loop sensing platform. This evolution requires unified systems that synergistically combine programmable excitation, high-speed computation, and intelligent feedback to drive its clinical adoption and enable precise diagnostics.

\section{Future work}
Although clinical applications of PAI have been extensively explored, and the International Photoacoustic Standardization Consortium has advocated for standardization in data management, hardware testing, and clinical protocols, PAI still faces significant barriers to routine clinical adoption. Despite receiving regulatory approval from the FDA and CE, as well as integration into the DICOM standard~\cite{7park2025clinical}, a critical evaluation reveals persistent technical and physical constraints.

First, laser safety is strictly governed by maximum permissible exposure (MPE) limits, such as those defined in ANSI Z136.1, which restrict optical fluence and consequently limit imaging depth, particularly in pigmented or highly absorbing tissues. Second, reliable acoustic coupling between transducers and human skin remains challenging owing to anatomical curvature, patient motion, hair, dependence on ultrasound gel, and factors that compromise sterility, hinder wearable deployment, and impede longitudinal monitoring. Third, despite advances in AI-accelerated reconstruction, current imaging speeds are often insufficient for real-time guidance in dynamic clinical scenarios, such as cardiac or interventional procedures.

Beyond these technical limitations, broader clinical and operational hurdles further impede translation. The high cost of PAI systems---driven by tunable lasers, dense ultrasound arrays, and hybrid imaging architectures---restricts their deployment to well-funded academic centers, limiting accessibility in point-of-care or resource-constrained settings. Moreover, the absence of universally accepted acquisition protocols and formalized operator training programs undermines reproducibility and diagnostic consistency across institutions and users. Although ongoing innovations in hardware, software, and artificial intelligence hold considerable promise, these intertwined practical, regulatory, and systemic challenges must be addressed systematically to enable broad clinical adoption.

Looking beyond current limitations, three transformative frontiers are poised to redefine the clinical impact of PAI. First, AI-enhanced PAI enables not only accelerated and artifact-robust reconstruction but also high-fidelity spectral unmixing of multiple contrast agents, which is critical for multiplexed molecular imaging and quantitative biomarker extraction. Second, the emergence of wearable and flexible PAI probes holds promise for continuous, noninvasive monitoring of hemodynamics, tissue oxygenation, and drug kinetics in ambulatory settings, although challenges in motion robustness and long-term acoustic coupling persist. Third, theranostic integration is advancing toward closed-loop interventions, in which PAI guides and monitors photothermal therapy or triggers photoacoustic-controlled drug release from stimuli-responsive nanocarriers in real time. Together, these directions bridge diagnostic imaging with personalized therapy, positioning PAI as an active enabler of precision medicine rather than merely a passive observation tool.

\subsection{Next-Generation Excitation Sources and System Miniaturization}
Recent developments in excitation technologies have driven the evolution of PAI toward more compact, efficient, and spectrally versatile systems. Although conventional OPOs and Q-switched lasers provide high pulse energies, their size, cost, and limited tunability have motivated the exploration of alternative sources, including NIR-II lasers, supercontinuum sources, and high-power LEDs or laser diodes. These emerging sources facilitate deeper tissue penetration, broadband tunability, and improved energy efficiency, enabling miniaturized, wearable, and point-of-care PAI implementation.

Furthermore, the combination of adaptive illumination and computational reconstruction, in which excitation is modulated in response to tissue feedback, represents a paradigm shift. This approach allows subacoustic-resolution imaging while maximizing the efficiency of compact light sources. Collectively, these advances have made PAI systems faster, smaller, and more accessible, accelerating their transition from experimental research to clinical applications.

\subsection{Single Image Super-Resolution in PAI}
High-resolution photoacoustic (PA) images are essential for clinical applications and image analysis; however, hardware limitations often necessitate trade-offs in scan time, SNR, and spatial coverage. DL-based super-resolution techniques, which have been extensively researched in other modalities, enhance PA image reconstruction through iterative optimization of the gradient flow, improve model stability, and capture detailed information. These techniques integrate dense connections and feedback mechanisms to leverage prior information for self-correction, leading to stable iterative optimization and accelerated model convergence, thus improving reconstruction fidelity.

Implementing super-resolution techniques in PA imaging enhances diagnostic value and analytical precision without requiring costly hardware upgrades. Advanced DL models tailored for super-resolution tasks enable higher-fidelity reconstruction, offering clearer visualization of anatomical structures and potential pathologies. Since the pioneering work of Dong \textit{et al.} in 2014 \cite{125dong2015image}, optimized models such as SRCNN \cite{126dong2014learning}, SRGAN \cite{127ledig2017photo}, CAMixerSR \cite{128wang2024camixersr}, and CoSeR \cite{129sun2024coser} have emerged, showing significant progress within the super-resolution technology domain that is specifically applicable across medical imaging contexts. Furthermore, deep-structured networks such as VDSR \cite{130kim2016accurate}, DRRN \cite{131tai2017image}, and RFDN \cite{132liu2020residual} combined with innovative frameworks exemplified by deep multiscale networks (DMSN) \cite{133wang2019transform} have further enhanced the quality metrics associated with existing SR methods utilized throughout healthcare settings.

\subsection{Physics-informed Reconstruction in PAI}
Using DL for PAI presents challenges, such as managing incomplete datasets and high computational costs associated with large training requirements. To address these issues, physics-informed DL approaches have been developed. These methods integrate physical models with AI to enhance reconstruction quality and reduce processing time. By incorporating physical constraints and prior knowledge, these approaches can improve the performance and robustness of DL architectures. Notable examples include PAT-MDAE \cite{75song2024accelerated} and a physics-driven DL-based filtered back-projection (dFBP) framework \cite{134shen2024physics}, which demonstrate significant potential for advancing biomedical PAI applications.

\subsection{DLearning-based Multimodal Fusion in PAI}
Multimodal imaging systems, combining PAI with OCT, ultrasound, and other modalities, enhance imaging depth and spatial resolution, providing comprehensive tissue information \cite{135zhu2023novel}. Multimodal image-guided surgical navigation, integrating techniques such as endoscopy, ultrasound, and fluorescence, improves visualization and spatial identification of critical structures, supporting minimally invasive procedures in neurosurgery, orthopedics, and vascular interventions. Key technologies include multimodal image segmentation, surgical planning, position calibration, image registration, and multi-source information fusion. Recent advancements in image fusion techniques, such as GANMcC \cite{136ma2020ganmcc}, leverage GANs alongside Transformers to effectively integrate gradient and intensity information. Nyayapathi \textit{et al.} \cite{137nyayapathi2024dual} proposed a novel architecture that integrates both Transformer and CNN elements, featuring global and local branches within the encoder module. Furthermore, the combination of photoacoustic/ultrasound systems with methods like diffuse optical tomography and MRI enhances disease monitoring and diagnosis; this is particularly beneficial for brain imaging aimed at obtaining functional and molecular insights through MRI. Anatomical data derived from ultrasound improves the lateral resolution of PAI systems, enabling detailed visualization of cerebral vasculature \cite{138notsuka2022improvement}. Additionally, it is crucial to evaluate these fusion technologies in clinical settings to ensure their continued advancement \cite{139biswas2016method}.

\subsection{Quantitative Analysis in PAI}
Future research on qPAI will focus on enhancing luminous flux modeling for precise tissue absorption coefficient reconstruction, improving spectral unmixing using advanced methods, and developing unsupervised super-resolution models to address data scarcity. Efforts will also include creating compact and efficient network models, innovating evaluation metrics aligned with human perception, and deepening DL theory for better reliability and interpretability. Advances in reconstruction techniques will address limited views and acoustic heterogeneity while enabling real-time reconstruction using DL. Integrating qPAI with other modalities improves resolution and contrast. Clinical trials will establish its efficacy and safety, and new contrast agents, portable systems, and functional/molecular imaging extensions will further advance qPAI in biomedical research and clinical diagnosis \cite{140le2022segmentation}.

\section{Conclusion}
While DL has substantially advanced PAI, improving reconstruction speed, spatial resolution, and quantitative accuracy, these advances entail inherent trade-offs. As highlighted in this review, challenges such as overfitting, limited training data, and poor model interpretability continue to hinder clinical translation. Moving forward, algorithmic development must prioritize not only performance but also robustness, transparency, and reproducibility. PAI has emerged as a powerful hybrid imaging modality, leveraging optical contrast and ultrasound resolution to overcome shallow penetration and strong scattering that limit purely optical techniques such as OCT. Over the past decade, PAI has shown considerable potential in disease screening, diagnosis, and longitudinal monitoring across diverse organ systems. This review outlines the fundamental principles of PAI and its advantages over conventional optical imaging and US, examines the limitations of traditional reconstruction and analysis methods, and emphasizes the transformative role of DL in advancing PACT, PAM, and PAE. AI-driven approaches have proven particularly effective in mitigating image degradation caused by sparse spatial sampling, which is a common constraint in practical PAI systems. By combining physics-informed architectures with data-driven priors, these methods reduce artifacts, improve structural delineation, enhance source localization, and enable reliable quantification of optical absorption. Nevertheless, two major challenges persist: compensating for wavelength- and depth-dependent optical fluence to obtain accurate absorption maps and achieving robust spectral unmixing for precise physiological and molecular characterization at depth. Addressing these challenges is essential to unlock the clinical potential of PAI. In summary, this review identifies several promising research directions, including single-image super-resolution, physics-informed DL, multimodal image fusion, and end-to-end quantitative biomarker analysis. Together, these pathways illustrate how DL continues to reshape and extend the capabilities of photoacoustic imaging, paving the way for broader clinical impact.

\backmatter



\section*{Declarations}

\bmhead*{Acknowledgments}
We thank the editors and anonymous reviewers for their valuable and constructive comments, which have greatly improved the quality of this manuscript.

\bmhead*{Authors’ contributions}
L.W. performed the original draft writing, methodology development, formal analysis, conceptualization, and software development.  
W.Z. reviewed and edited the manuscript, supervised the project, and was responsible for project administration and funding acquisition.  
K.L. and R.L. contributed to revising the manuscript; H.C. assisted with graphic design and manuscript review.  
L.L. provided support in data collection and preprocessing.  
W.T.S. contributed to language editing and manuscript formatting.  
N.W. reviewed and edited the manuscript, supervised the project, administered both the project and funding acquisition, and contributed to conceptualization and methodology.  
All authors reviewed and edited the manuscript.

\bmhead*{Funding}
This work was supported in part by the Hong Kong Polytechnic University Start-up Fund (Project ID: P0053210), the Hong Kong Polytechnic University Departmental Collaborative Research Fund (Project ID: P0056428), and the Hong Kong RGC General Research Fund (GRF) under Grants 14220622 and 14204321.

\bmhead*{Availability of data and materials}
Not applicable.

\bmhead*{Competing interests}
The authors declare that they have no competing interests.



\nocite{*}

\bibliography{sn-bibliography}

@inproceedings{1oraevsky1994laser,
  author    = "Oraevsky, A. A. and Jacques, S. L. and Esenaliev, R. O. and Tittel, F. K.",
  title     = "Laser-based optoacoustic imaging in biological tissues",
  booktitle = "In: Laser-Tissue Interaction V; and Ultraviolet Radiation Hazards",
  volume    = "2134",
  pages     = "122--128",
  year      = "1994",
  address   = "San Jose, CA",
  publisher = "SPIE"
}

@article{2matthews2017joint,
  author    = "Matthews, T. P. and Anastasio, M. A.",
  title     = "Joint reconstruction of the initial pressure and speed of sound distributions from combined photoacoustic and ultrasound tomography measurements",
  journal   = "Inverse Probl.",
  volume    = "33",
  number    = "12",
  pages     = "124002",
  year      = "2017",
  doi       = "10.1088/1361-6420/aa93fd"
}

@article{3wang2008tutorial,
  author    = "Wang, L. V.",
  title     = "Tutorial on photoacoustic microscopy and computed tomography",
  journal   = "IEEE J. Sel. Top. Quantum Electron.",
  volume    = "14",
  number    = "1",
  pages     = "171--179",
  year      = "2008",
  doi       = "10.1109/JSTQE.2007.913523"
}

@article{4poudel2019survey,
  author    = "Poudel, J. and Lou, Y. and Anastasio, M. A.",
  title     = "A survey of computational frameworks for solving the acoustic inverse problem in three-dimensional photoacoustic computed tomography",
  journal   = "Phys. Med. Biol.",
  volume    = "64",
  number    = "14",
  pages     = "14TR01",
  year      = "2019",
  doi       = "10.1088/1361-6560/ab166f"
}

@article{5ning2015ultrasound,
  author    = "Ning, B. and Sun, N. and Cao, R. and Chen, R. and Shung, K. K. and Hossack, J. A. and Lee, J.-M. and Zhou, Q. and Hu, S.",
  title     = "Ultrasound-aided multi-parametric photoacoustic microscopy of the mouse brain",
  journal   = "Sci. Rep.",
  volume    = "5",
  number    = "1",
  pages     = "18775",
  year      = "2015",
  doi       = "10.1038/srep18775"
}

@article{6zackrisson2014light,
  author    = "Zackrisson, S. and Van De Ven, S. M. W. Y. and Gambhir, S. S.",
  title     = "Light in and sound out: emerging translational strategies for photoacoustic imaging",
  journal   = "Cancer Res.",
  volume    = "74",
  number    = "4",
  pages     = "979--1004",
  year      = "2014",
  doi       = "10.1158/0008-5472.CAN-13-2388"
}

@article{7park2025clinical,
  author    = "Park, J. and Choi, S. and Knieling, F. and Clingman, B. and Bohndiek, S. and Wang, L. V. and Kim, C.",
  title     = "Clinical translation of photoacoustic imaging",
  journal   = "Nat. Rev. Bioeng.",
  volume    = "3",
  number    = "3",
  pages     = "193--212",
  year      = "2025",
  doi       = "10.1038/s44222-025-00070-8"
}

@article{8park2017contrast,
  author    = "Park, S. and Jung, U. and Lee, S. and Lee, D. and Kim, C.",
  title     = "Contrast-enhanced dual mode imaging: photoacoustic imaging plus more",
  journal   = "Biomed. Eng. Lett.",
  volume    = "7",
  number    = "2",
  pages     = "121--133",
  year      = "2017",
  doi       = "10.1007/s13534-017-0024-9"
}

@article{9rajendran2022photoacoustic,
  author    = "Rajendran, P. and Sharma, A. and Pramanik, M.",
  title     = "Photoacoustic imaging aided with deep learning: a review",
  journal   = "Biomed. Eng. Lett.",
  volume    = "12",
  number    = "2",
  pages     = "155--173",
  year      = "2022",
  doi       = "10.1007/s13534-022-00220-5"
}

@article{10dispirito2021sounding,
  author    = "DiSpirito, A. III and Vu, T. and Pramanik, M. and Yao, J.",
  title     = "Sounding out the hidden data: a concise review of deep learning in photoacoustic imaging",
  journal   = "Exp. Biol. Med.",
  volume    = "246",
  number    = "12",
  pages     = "1355--1367",
  year      = "2021",
  doi       = "10.1177/15353702211012786"
}

@article{13guo2020photoacoustic,
  author    = "Guo, H. and Li, Y. and Qi, W. and Xi, L.",
  title     = "Photoacoustic endoscopy: A progress review",
  journal   = "J. Biophotonics",
  volume    = "13",
  number    = "12",
  pages     = "e202000217",
  year      = "2020",
  doi       = "10.1002/jbio.202000217"
}

@article{14cox2012quantitative,
  author    = "Cox, B. and Laufer, J. G. and Arridge, S. R. and Beard, P. C.",
  title     = "Quantitative spectroscopic photoacoustic imaging: a review",
  journal   = "J. Biomed. Opt.",
  volume    = "17",
  number    = "6",
  pages     = "061202",
  year      = "2012",
  doi       = "10.1117/1.JBO.17.6.061202"
}

@article{16park2017real,
  author    = "Park, S. and Jang, J. and Kim, J. and Kim, Y. S. and Kim, C.",
  title     = "Real-time triple-modal photoacoustic, ultrasound, and magnetic resonance fusion imaging of humans",
  journal   = "IEEE Trans. Med. Imaging",
  volume    = "36",
  number    = "9",
  pages     = "1912--1921",
  year      = "2017",
  doi       = "10.1109/TMI.2017.2698438"
}

@article{17mallidi2011photoacoustic,
  author    = "Mallidi, S. and Luke, G. P. and Emelianov, S.",
  title     = "Photoacoustic imaging in cancer detection, diagnosis, and treatment guidance",
  journal   = "Trends Biotechnol.",
  volume    = "29",
  number    = "5",
  pages     = "213--221",
  year      = "2011",
  doi       = "10.1016/j.tibtech.2011.01.007"
}

@article{18beard2011biomedical,
  author    = "Beard, P.",
  title     = "Biomedical photoacoustic imaging",
  journal   = "Interface Focus",
  volume    = "1",
  number    = "4",
  pages     = "602--631",
  year      = "2011",
  doi       = "10.1098/rsfs.2011.0028"
}

@article{19yu2024simultaneous,
  author    = "Yu, Y. and Feng, T. and Qiu, H. and Gu, Y. and Chen, Q. and Zuo, C. and Ma, H.",
  title     = "Simultaneous photoacoustic and ultrasound imaging: A review",
  journal   = "Ultrasonics",
  volume    = "139",
  pages     = "107277",
  year      = "2024",
  doi       = "10.1016/j.ultras.2024.107277"
}

@article{20wang2009multiscale,
  author    = "Wang, L. V.",
  title     = "Multiscale photoacoustic microscopy and computed tomography",
  journal   = "Nat. Photonics",
  volume    = "3",
  number    = "9",
  pages     = "503--509",
  year      = "2009",
  doi       = "10.1038/nphoton.2009.157"
}

@article{21liu2019integrated,
  author    = "Liu, C. and Liao, J. and Chen, L. and Chen, J. and Ding, R. and Gong, X. and Cui, C. and Pang, Z. and Zheng, W. and Song, L.",
  title     = "The integrated high-resolution reflection-mode photoacoustic and fluorescence confocal microscopy",
  journal   = "Photoacoustics",
  volume    = "14",
  pages     = "12--18",
  year      = "2019",
  doi       = "10.1016/j.pacs.2019.03.002"
}

@article{22attia2019review,
  author    = "Attia, A. B. E. and Balasundaram, G. and Moothanchery, M. and Dinish, U. S. and Bi, R. and Ntziachristos, V. and Olivo, M.",
  title     = "A review of clinical photoacoustic imaging: Current and future trends",
  journal   = "Photoacoustics",
  volume    = "16",
  pages     = "100144",
  year      = "2019",
  doi       = "10.1016/j.pacs.2019.100144"
}

@article{23omar2019optoacoustic,
  author    = "Omar, M. and Aguirre, J. and Ntziachristos, V.",
  title     = "Optoacoustic mesoscopy for biomedicine",
  journal   = "Nat. Biomed. Eng.",
  volume    = "3",
  number    = "5",
  pages     = "354--370",
  year      = "2019",
  doi       = "10.1038/s41551-019-0379-6"
}

@article{24moothanchery2017performance,
  author    = "Moothanchery, M. and Pramanik, M.",
  title     = "Performance characterization of a switchable acoustic resolution and optical resolution photoacoustic microscopy system",
  journal   = "Sensors",
  volume    = "17",
  number    = "2",
  pages     = "357",
  year      = "2017",
  doi       = "10.3390/s17020357"
}

@article{25lin2022emerging,
  author    = "Lin, L. and Wang, L. V.",
  title     = "The emerging role of photoacoustic imaging in clinical oncology",
  journal   = "Nat. Rev. Clin. Oncol.",
  volume    = "19",
  number    = "6",
  pages     = "365--384",
  year      = "2022",
  doi       = "10.1038/s41571-022-00623-3"
}

@article{26_2021Recent,
  author    = "Wang, L. V. and others",
  title     = "Recent advances in photoacoustic tomography",
  journal   = "BME Front.",
  year      = "2021",
  doi       = "10.34133/2021/9857438"
}

@article{27na2021photoacoustic,
  author    = "Na, S. and Wang, L. V.",
  title     = "Photoacoustic computed tomography for functional human brain imaging",
  journal   = "Biomed. Opt. Express",
  volume    = "12",
  number    = "7",
  pages     = "4056--4083",
  year      = "2021",
  doi       = "10.1364/BOE.425863"
}

@article{28bauer2011quantitative,
  author    = "Bauer, A. Q. and Nothdurft, R. E. and Erpelding, T. N. and Wang, L. V. and Culver, J. P.",
  title     = "Quantitative photoacoustic imaging: correcting for heterogeneous light fluence distributions using diffuse optical tomography",
  journal   = "J. Biomed. Opt.",
  volume    = "16",
  number    = "9",
  pages     = "096016",
  year      = "2011",
  doi       = "10.1117/1.3615567"
}

@article{29tian2021spatial,
  author    = "Tian, C. and Zhang, C. and Zhang, H. and Xie, D. and Jin, Y.",
  title     = "Spatial resolution in photoacoustic computed tomography",
  journal   = "Rep. Prog. Phys.",
  volume    = "84",
  number    = "3",
  pages     = "036701",
  year      = "2021",
  doi       = "10.1088/1361-6633/abc9a7"
}

@article{30jeon2013multimodal,
  author    = "Jeon, M. and Kim, C.",
  title     = "Multimodal photoacoustic tomography",
  journal   = "IEEE Trans. Multimed.",
  volume    = "15",
  number    = "5",
  pages     = "975--982",
  year      = "2013",
  doi       = "10.1109/TMM.2013.2253771"
}

@article{31jeon2019review,
  author    = "Jeon, S. and Kim, J. and Lee, D. and Baik, J. W. and Kim, C.",
  title     = "Review on practical photoacoustic microscopy",
  journal   = "Photoacoustics",
  volume    = "15",
  pages     = "100141",
  year      = "2019",
  doi       = "10.1016/j.pacs.2019.100141"
}

@article{32vu2020generative,
  author    = "Vu, T. and Li, M. and Humayun, H. and Zhou, Y. and Yao, J.",
  title     = "A generative adversarial network for artifact removal in photoacoustic computed tomography with a linear-array transducer",
  journal   = "Exp. Biol. Med.",
  volume    = "245",
  number    = "7",
  pages     = "597--605",
  year      = "2020",
  doi       = "10.1177/1535370220914289"
}

@article{33steinberg2019photoacoustic,
  author    = "Steinberg, I. and Huland, D. M. and Vermesh, O. and Frostig, H. E. and Tummers, W. S. and Gambhir, S. S.",
  title     = "Photoacoustic clinical imaging",
  journal   = "Photoacoustics",
  volume    = "14",
  pages     = "77--98",
  year      = "2019",
  doi       = "10.1016/j.pacs.2019.05.001"
}

@article{34rajendran2020deep,
  author    = "Rajendran, P. and Pramanik, M.",
  title     = "Deep learning approach to improve tangential resolution in photoacoustic tomography",
  journal   = "Biomed. Opt. Express",
  volume    = "11",
  number    = "12",
  pages     = "7311--7323",
  year      = "2020",
  doi       = "10.1364/BOE.408327"
}

@article{35zhang2024challenges,
  author    = "Zhang, S. and Miao, J. and Li, L. S.",
  title     = "Challenges and advances in two-dimensional photoacoustic computed tomography: a review",
  journal   = "J. Biomed. Opt.",
  volume    = "29",
  number    = "7",
  pages     = "070901",
  year      = "2024",
  doi       = "10.1117/1.JBO.29.7.070901"
}

@article{36tian2020impact,
  author    = "Tian, C. and Pei, M. and Shen, K. and Liu, S. and Hu, Z. and Feng, T.",
  title     = "Impact of system factors on the performance of photoacoustic tomography scanners",
  journal   = "Phys. Rev. Appl.",
  volume    = "13",
  number    = "1",
  pages     = "014001",
  year      = "2020",
  doi       = "10.1103/PhysRevApplied.13.014001"
}

@article{37yao2021perspective,
  author    = "Yao, J. and Wang, L. V.",
  title     = "Perspective on fast-evolving photoacoustic tomography",
  journal   = "J. Biomed. Opt.",
  volume    = "26",
  number    = "6",
  pages     = "060602",
  year      = "2021",
  doi       = "10.1117/1.JBO.26.6.060602"
}

@article{38zhou2016tutorial,
  author    = "Zhou, Y. and Yao, J. and Wang, L. V.",
  title     = "Tutorial on photoacoustic tomography",
  journal   = "J. Biomed. Opt.",
  volume    = "21",
  number    = "6",
  pages     = "061007",
  year      = "2016",
  doi       = "10.1117/1.JBO.21.6.061007"
}

@article{39balci2025enhanced,
  author    = "Balci, Z. and Mert, A.",
  title     = "Enhanced photoacoustic signal processing using empirical mode decomposition and machine learning",
  journal   = "Nondestruct. Test. Eval.",
  volume    = "40",
  number    = "5",
  pages     = "2044--2056",
  year      = "2025",
  doi       = "10.1080/10589759.2024.2423456"
}

@article{40yao2013photoacoustic,
  author    = "Yao, J. and Wang, L. V.",
  title     = "Photoacoustic microscopy",
  journal   = "Laser Photonics Rev.",
  volume    = "7",
  number    = "5",
  pages     = "758--778",
  year      = "2013",
  doi       = "10.1002/lpor.201200060"
}

@article{41kempski2020application,
  author    = "Kempski, K. M. and Graham, M. T. and Gubbi, M. R. and Palmer, T. and Lediju Bell, M. A.",
  title     = "Application of the generalized contrast-to-noise ratio to assess photoacoustic image quality",
  journal   = "Biomed. Opt. Express",
  volume    = "11",
  number    = "7",
  pages     = "3684--3698",
  year      = "2020",
  doi       = "10.1364/BOE.392779"
}

@article{42chen2021progress,
  author    = "Chen, Q. and Qin, W. and Qi, W. and Xi, L.",
  title     = "Progress of clinical translation of handheld and semi-handheld photoacoustic imaging",
  journal   = "Photoacoustics",
  volume    = "22",
  pages     = "100264",
  year      = "2021",
  doi       = "10.1016/j.pacs.2021.100264"
}

@article{43hsu2021comparing,
  author    = "Hsu, K.-T. and Guan, S. and Chitnis, P. V.",
  title     = "Comparing deep learning frameworks for photoacoustic tomography image reconstruction",
  journal   = "Photoacoustics",
  volume    = "23",
  pages     = "100271",
  year      = "2021",
  doi       = "10.1016/j.pacs.2021.100271"
}

@article{44yoon2013recent,
  author    = "Yoon, T.-J. and Cho, Y.-S.",
  title     = "Recent advances in photoacoustic endoscopy",
  journal   = "World J. Gastrointest. Endosc.",
  volume    = "5",
  number    = "11",
  pages     = "534",
  year      = "2013",
  doi       = "10.4253/wjge.v5.i11.534"
}

@article{45Seong2020,
  author    = "Seong, M. and Chen, S.-L.",
  title     = "Recent advances toward clinical applications of photoacoustic microscopy: a review",
  journal   = "Sci. China Life Sci.",
  volume    = "63",
  number    = "12",
  pages     = "1798--1812",
  year      = "2020",
  doi       = "10.1007/s11427-020-1758-1"
}

@article{46_2024Image,
  author    = "Tian, C. and Shen, K. and Dong, W. and Gao, F. and Wang, K. and Li, J. and Liu, S. and Feng, T. and Liu, C. and Li, C. and others",
  title     = "Image reconstruction from photoacoustic projections",
  journal   = "Photonics Insights",
  volume    = "3",
  number    = "3",
  pages     = "R06",
  year      = "2024",
  doi       = "10.1117/1.PHI.3.3.R06"
}

@article{47xu2004time,
  author    = "Xu, Y. and Wang, L. V.",
  title     = "Time reversal and its application to tomography with diffracting sources",
  journal   = "Phys. Rev. Lett.",
  volume    = "92",
  number    = "3",
  pages     = "033902",
  year      = "2004",
  doi       = "10.1103/PhysRevLett.92.033902"
}

@article{48burgholzer2007exact,
  author    = "Burgholzer, P. and Matt, G. J. and Haltmeier, M. and Paltauf, G{\"u}nther",
  title     = "Exact and approximative imaging methods for photoacoustic tomography using an arbitrary detection surface",
  journal   = "Phys. Rev. E",
  volume    = "75",
  number    = "4",
  pages     = "046706",
  year      = "2007",
  doi       = "10.1103/PhysRevE.75.046706"
}

@article{49hoelen1998three,
  author    = "Hoelen, C. G. A. and De Mul, F. F. M. and Pongers, R. and Dekker, A.",
  title     = "Three-dimensional photoacoustic imaging of blood vessels in tissue",
  journal   = "Opt. Lett.",
  volume    = "23",
  number    = "8",
  pages     = "648--650",
  year      = "1998",
  doi       = "10.1364/OL.23.000648"
}

@inproceedings{50hoelen1998photoacoustic,
  author    = "Hoelen, C. G. A. and Pongers, R. and Hamhuis, G. and de Mul, F. F. M. and Greve, J.",
  title     = "Photoacoustic blood cell detection and imaging of blood vessels in phantom tissue",
  booktitle = "Optical and Imaging Techniques for Biomonitoring III",
  series    = "Proceedings of {SPIE}",
  volume    = "3196",
  pages     = "142--153",
  year      = "1998",
  address   = "The Hague, Netherlands",
  publisher = "SPIE",
  doi       = "10.1117/12.319422"
}

@article{51kruger1995photoacoustic,
  author    = "Kruger, R. A. and Liu, P. and Fang, Y. ``Richard'' and Appledorn, C. R.",
  title     = "Photoacoustic ultrasound (PAUS)—reconstruction tomography",
  journal   = "Med. Phys.",
  volume    = "22",
  number    = "10",
  pages     = "1605--1609",
  year      = "1995",
  doi       = "10.1118/1.597422"
}

@article{52guo2016deep,
  author    = "Guo, Y. and Liu, Y. and Oerlemans, A. and Lao, S. and Wu, S. and Lew, M. S.",
  title     = "Deep learning for visual understanding: A review",
  journal   = "Neurocomputing",
  volume    = "187",
  pages     = "27--48",
  year      = "2016",
  doi       = "10.1016/j.neucom.2015.12.088"
}

@article{53xu2005universal,
  author    = "Xu, M. and Wang, L. V.",
  title     = "Universal back-projection algorithm for photoacoustic computed tomography",
  journal   = "Phys. Rev. E",
  volume    = "71",
  number    = "1",
  pages     = "016706",
  year      = "2005",
  doi       = "10.1103/PhysRevE.71.016706"
}

@article{54tai2009se,
  author    = "Tai, C.-H. and Vincent, J. J. and Kim, C. and Lee, B.",
  title     = "SE: an algorithm for deriving sequence alignment from a pair of superimposed structures",
  journal   = "BMC Bioinformatics",
  volume    = "10",
  number    = "Suppl 1",
  pages     = "S4",
  year      = "2009",
  doi       = "10.1186/1471-2105-10-S1-S4"
}

@article{55wang2023photoacoustic,
  author    = "Wang, R. and Zhu, J. and Xia, J. and Yao, J. and Shi, J. and Li, C.",
  title     = "Photoacoustic imaging with limited sampling: a review of machine learning approaches",
  journal   = "Biomed. Opt. Express",
  volume    = "14",
  number    = "4",
  pages     = "1777--1799",
  year      = "2023",
  doi       = "10.1364/BOE.485654"
}

@inproceedings{56shan2019simultaneous,
  author    = "Shan, H. and Wiedeman, C. and Wang, G. and Yang, Y.",
  title     = "Simultaneous reconstruction of the initial pressure and sound speed in photoacoustic tomography using a deep-learning approach",
  booktitle = "Novel Optical Systems, Methods, and Applications {XXII}",
  series    = "Proceedings of {SPIE}",
  volume    = "11105",
  pages     = "18--27",
  year      = "2019",
  address   = "San Diego, California, United States",
  publisher = "SPIE",
  doi       = "10.1117/12.2529307"
}

@article{57ravishankar2019image,
  author    = "Ravishankar, S. and Ye, J. C. and Fessler, J. A.",
  title     = "Image reconstruction: From sparsity to data-adaptive methods and machine learning",
  journal   = "Proc. IEEE",
  volume    = "108",
  number    = "1",
  pages     = "86--109",
  year      = "2019",
  doi       = "10.1109/JPROC.2019.2941752"
}

@article{58davoudi2019deep,
  author    = "Davoudi, N. and De{\'a}n-Ben, X. L. and Razansky, D.",
  title     = "Deep learning optoacoustic tomography with sparse data",
  journal   = "Nat. Mach. Intell.",
  volume    = "1",
  number    = "10",
  pages     = "453--460",
  year      = "2019",
  doi       = "10.1038/s42256-019-0105-7"
}

@inproceedings{59antholzer2018photoacoustic,
  author    = "Antholzer, S. and Haltmeier, M. and Nuster, R. and Schwab, J.",
  title     = "Photoacoustic image reconstruction via deep learning",
  booktitle = "Photons Plus Ultrasound: Imaging and Sensing 2018",
  series    = "Proceedings of {SPIE}",
  volume    = "10494",
  pages     = "433--442",
  year      = "2018",
  address   = "San Francisco, California, United States",
  publisher = "SPIE",
  doi       = "10.1117/12.2288304"
}

@article{60wei2024deep,
  author    = "Wei, X. and Feng, T. and Huang, Q. and Chen, Q. and Zuo, C. and Ma, H.",
  title     = "Deep learning-powered biomedical photoacoustic imaging",
  journal   = "Neurocomputing",
  volume    = "573",
  pages     = "127207",
  year      = "2024",
  doi       = "10.1016/j.neucom.2023.127207"
}

@article{61yang2021review,
  author    = "Yang, C. and Lan, H. and Gao, F. and Gao, F.",
  title     = "Review of deep learning for photoacoustic imaging",
  journal   = "Photoacoustics",
  volume    = "21",
  pages     = "100215",
  year      = "2021",
  doi       = "10.1016/j.pacs.2021.100215"
}

@article{62huang2023review,
  author    = "Huang, Q. and Tian, H. and Jia, L. and Li, Z. and Zhou, Z.",
  title     = "A review of deep learning segmentation methods for carotid artery ultrasound images",
  journal   = "Neurocomputing",
  volume    = "545",
  pages     = "126298",
  year      = "2023",
  doi       = "10.1016/j.neucom.2022.126298"
}

@article{63schmidhuber2015deep,
  title={Deep learning in neural networks: An overview},
  author={Schmidhuber, J{\"u}rgen},
  journal={Neural networks},
  volume={61},
  pages={85--117},
  year={2015},
  publisher={Elsevier}
}

@inproceedings{64ronneberger2015u,
  author    = "Ronneberger, O. and Fischer, P. and Brox, T.",
  title     = "{U-Net}: Convolutional Networks for Biomedical Image Segmentation",
  booktitle = "Medical Image Computing and Computer-Assisted Intervention -- {MICCAI} 2015",
  series    = "Lecture Notes in Computer Science",
  volume    = "9351",
  pages     = "234--241",
  year      = "2015",
  address   = "Munich, Germany",
  publisher = "Springer",
  doi       = "10.1007/978-3-319-24574-4_28"
}

@article{65allman2018photoacoustic,
  author    = "Allman, D. and Reiter, A. and Bell, M. A. L.",
  title     = "Photoacoustic source detection and reflection artifact removal enabled by deep learning",
  journal   = "IEEE Trans. Med. Imaging",
  volume    = "37",
  number    = "6",
  pages     = "1464--1477",
  year      = "2018",
  doi       = "10.1109/TMI.2018.2828883"
}

@article{66gutta2017deep,
  author    = "Gutta, S. and Kadimesetty, V. S. and Kalva, S. K. and Pramanik, M. and Ganapathy, S. and Yalavarthy, P. K.",
  title     = "Deep neural network-based bandwidth enhancement of photoacoustic data",
  journal   = "J. Biomed. Opt.",
  volume    = "22",
  number    = "11",
  pages     = "116001",
  year      = "2017",
  doi       = "10.1117/1.JBO.22.11.116001"
}

@article{67awasthi2020deep,
  author    = "Awasthi, N. and Jain, G. and Kalva, S. K. and Pramanik, M. and Yalavarthy, P. K.",
  title     = "Deep neural network-based sinogram super-resolution and bandwidth enhancement for limited-data photoacoustic tomography",
  journal   = "IEEE Trans. Ultrason. Ferroelectr. Freq. Control",
  volume    = "67",
  number    = "12",
  pages     = "2660--2673",
  year      = "2020",
  doi       = "10.1109/TUFFC.2020.3018560"
}

@inproceedings{68durairaj2020unsupervised,
  author    = "Durairaj, D. A. and Agrawal, S. and Johnstonbaugh, K. and Chen, H. and Karri, S. P. K. and Kothapalli, S.-R.",
  title     = "Unsupervised deep learning approach for photoacoustic spectral unmixing",
  booktitle = "Photons Plus Ultrasound: Imaging and Sensing 2020",
  series    = "Proceedings of {SPIE}",
  volume    = "11240",
  pages     = "173--181",
  year      = "2020",
  address   = "San Francisco, California, United States",
  publisher = "SPIE",
  doi       = "10.1117/12.2546210"
}

@article{69feng2020end,
  author    = "Feng, J. and Deng, J. and Li, Z. and Sun, Z. and Dou, H. and Jia, K.",
  title     = "End-to-end Res-Unet based reconstruction algorithm for photoacoustic imaging",
  journal   = "Biomed. Opt. Express",
  volume    = "11",
  number    = "9",
  pages     = "5321--5340",
  year      = "2020",
  doi       = "10.1364/BOE.399954"
}

@article{70guan2019fully,
  author    = "Guan, S. and Khan, A. A. and Sikdar, S. and Chitnis, P. V.",
  title     = "Fully dense UNet for 2-D sparse photoacoustic tomography artifact removal",
  journal   = "IEEE J. Biomed. Health Inform.",
  volume    = "24",
  number    = "2",
  pages     = "568--576",
  year      = "2019",
  doi       = "10.1109/JBHI.2019.2930943"
}

@inproceedings{71zhang2021limited,
  author    = "Zhang, J. and Lan, H. and Yang, C. and Lyu, T. and Guo, S. and Gao, F. and Gao, F.",
  title     = "Limited-view photoacoustic imaging reconstruction with dual domain inputs based on mutual information",
  booktitle = "2021 {IEEE} 18th International Symposium on Biomedical Imaging ({ISBI})",
  pages     = "1522--1526",
  year      = "2021",
  address   = "Nice, France",
  publisher = "IEEE",
  doi       = "10.1109/ISBI48211.2021.9434158"
}

@article{72lan2023jointed,
  author    = "Lan, H. and Yang, C. and Gao, F.",
  title     = "A jointed feature fusion framework for photoacoustic image reconstruction",
  journal   = "Photoacoustics",
  volume    = "29",
  pages     = "100442",
  year      = "2023",
  doi       = "10.1016/j.pacs.2023.100442"
}

@article{73zhang2019progressive,
  author    = "Zhang, D. and Khoreva, A.",
  title     = "Progressive augmentation of GANs",
  journal   = "Adv. Neural Inf. Process. Syst.",
  volume    = "32",
  year      = "2019",
  doi       = "10.48550/arXiv.1906.05092"
}

@article{74zhong2024unsupervised,
  author    = "Zhong, W. and Li, T. and Hou, S. and Zhang, H. and Li, Z. and Wang, G. and Liu, Q. and Song, X.",
  title     = "Unsupervised disentanglement strategy for mitigating artifact in photoacoustic tomography under extremely sparse view",
  journal   = "Photoacoustics",
  volume    = "38",
  pages     = "100613",
  year      = "2024",
  doi       = "10.1016/j.pacs.2024.100613"
}

@article{75song2024accelerated,
  author    = "Song, X. and Zhong, W. and Li, Z. and Peng, S. and Zhang, H. and Wang, G. and Dong, J. and Liu, X. and Xu, X. and Liu, Q.",
  title     = "Accelerated model-based iterative reconstruction strategy for sparse-view photoacoustic tomography aided by multi-channel autoencoder priors",
  journal   = "J. Biophotonics",
  volume    = "17",
  number    = "1",
  pages     = "e202300281",
  year      = "2024",
  doi       = "10.1002/jbio.202300281"
}

@article{76guo2022net,
  author    = "Guo, M. and Lan, H. and Yang, C. and Liu, J. and Gao, F.",
  title     = "AS-Net: fast photoacoustic reconstruction with multi-feature fusion from sparse data",
  journal   = "IEEE Trans. Comput. Imaging",
  volume    = "8",
  pages     = "215--223",
  year      = "2022",
  doi       = "10.1109/TCI.2022.3141592"
}

@inproceedings{77chan2024local,
  author    = "Chan, S. C. K. and Shi, L. and Huang, B. and Wong, T. T. W.",
  title     = "Local Spatial Attention Transformer for Sparse Photoacoustic Image Reconstruction",
  booktitle = "2024 {IEEE} International Symposium on Biomedical Imaging ({ISBI})",
  pages     = "1--5",
  year      = "2024",
  address   = "Athens, Greece",
  publisher = "IEEE",
  doi       = "10.1109/ISBI57699.2024.10530001"
}

@article{78tong2020domain,
  author    = "Tong, T. and Huang, W. and Wang, K. and He, Z. and Yin, L. and Yang, X. and Zhang, S. and Tian, J.",
  title     = "Domain transform network for photoacoustic tomography from limited-view and sparsely sampled data",
  journal   = "Photoacoustics",
  volume    = "19",
  pages     = "100190",
  year      = "2020",
  doi       = "10.1016/j.pacs.2020.100190"
}

@article{79lan2020net,
  author    = "Lan, H. and Jiang, D. and Yang, C. and Gao, F. and Gao, F.",
  title     = "Y-Net: Hybrid deep learning image reconstruction for photoacoustic tomography in vivo",
  journal   = "Photoacoustics",
  volume    = "20",
  pages     = "100197",
  year      = "2020",
  doi       = "10.1016/j.pacs.2020.100197"
}

@article{80vu2021deep,
  author    = "Vu, T. and DiSpirito III, A. and Li, D. and Wang, Z. and Zhu, X. and Chen, M. and Jiang, L. and Zhang, D. and Luo, J. and Zhang, Y. S. and others",
  title     = "Deep image prior for undersampling high-speed photoacoustic microscopy",
  journal   = "Photoacoustics",
  volume    = "22",
  pages     = "100266",
  year      = "2021",
  doi       = "10.1016/j.pacs.2021.100266"
}

@article{81seong2023three,
  author    = "Seong, D. and Lee, E. and Kim, Y. and Han, S. and Lee, J. and Jeon, M. and Kim, J.",
  title     = "Three-dimensional reconstructing undersampled photoacoustic microscopy images using deep learning",
  journal   = "Photoacoustics",
  volume    = "29",
  pages     = "100429",
  year      = "2023",
  doi       = "10.1016/j.pacs.2022.100429"
}

@article{82li2024removing,
  author    = "Li, B. and Lu, M. and Zhou, T. and Bu, M. and Gu, W. and Wang, J. and Zhu, Q. and Liu, X. and Ta, D.",
  title     = "Removing Artifacts in Transcranial Photoacoustic Imaging With Polarized Self-Attention Dense-UNet",
  journal   = "Ultrasound Med. Biol.",
  volume    = "50",
  number    = "10",
  pages     = "1530--1543",
  year      = "2024",
  doi       = "10.1016/j.ultrasmedbio.2024.05.007"
}

@article{83wang2024reconstructing,
  author    = "Wang, J. and Li, B. and Zhou, T. and Liu, C. and Lu, M. and Gu, W. and Liu, X. and Ta, D.",
  title     = "Reconstructing cancellous bone from down-sampled optical-resolution photoacoustic microscopy images with deep learning",
  journal   = "Ultrasound Med. Biol.",
  volume    = "50",
  number    = "9",
  pages     = "1459--1471",
  year      = "2024",
  doi       = "10.1016/j.ultrasmedbio.2024.04.012"
}

@inproceedings{84shahid2021batch,
  author    = "Shahid, H. and Yue, Y. and Khalid, A. and Liu, X. and Ta, D.",
  title     = "Batch renormalization accumulated residual U-network for artifacts removal in photoacoustic imaging",
  booktitle = "2021 {IEEE} International Ultrasonics Symposium ({IUS})",
  pages     = "1--4",
  year      = "2021",
  address   = "Xi'an, China",
  publisher = "IEEE",
  doi       = "10.1109/IUS52206.2021.9593428"
}

@article{85cao2025mean,
  author    = "Cao, Y. and Lu, S. and Wan, C. and Wang, Y. and Liu, X. and Guo, K. and Cao, Y. and Li, Z. and Liu, Q. and Song, X.",
  title     = "Mean-reverting diffusion model-enhanced acoustic-resolution photoacoustic microscopy for resolution enhancement: Toward optical resolution",
  journal   = "J. Innov. Opt. Health Sci.",
  volume    = "18",
  number    = "02",
  pages     = "2450023",
  year      = "2025",
  doi       = "10.1142/S1793545824500235"
}

@article{86le2023enhanced,
  author    = "Le, T. D. and Min, J.-J. and Lee, C.",
  title     = "Enhanced resolution and sensitivity acoustic-resolution photoacoustic microscopy with semi/unsupervised GANs",
  journal   = "Sci. Rep.",
  volume    = "13",
  number    = "1",
  pages     = "13423",
  year      = "2023",
  doi       = "10.1038/s41598-023-39990-5"
}

@article{87cheng2022high,
  author    = "Cheng, S. and Zhou, Y. and Chen, J. and Li, H. and Wang, L. and Lai, P.",
  title     = "High-resolution photoacoustic microscopy with deep penetration through learning",
  journal   = "Photoacoustics",
  volume    = "25",
  pages     = "100314",
  year      = "2022",
  doi       = "10.1016/j.pacs.2021.100314"
}

@article{88zhao2021deep,
  author    = "Zhao, H. and Ke, Z. and Yang, F. and Li, K. and Chen, N. and Song, L. and Zheng, C. and Liang, D. and Liu, C.",
  title     = "Deep learning enables superior photoacoustic imaging at ultralow laser dosages",
  journal   = "Adv. Sci.",
  volume    = "8",
  number    = "3",
  pages     = "2003097",
  year      = "2021",
  doi       = "10.1002/advs.202003097"
}

@article{89zhou2019noise,
  author    = "Zhou, M. and Xia, H. and Zhong, H. and Zhang, J. and Gao, F.",
  title     = "A noise reduction method for photoacoustic imaging in vivo based on EMD and conditional mutual information",
  journal   = "IEEE Photonics J.",
  volume    = "11",
  number    = "1",
  pages     = "1--10",
  year      = "2019",
  doi       = "10.1109/JPHOT.2019.2891398"
}

@article{90liu2024upamnet,
  author    = "Liu, Y. and Zhou, J. and Luo, Y. and Li, J. and Chen, S.-L. and Guo, Y. and Yang, G.-Z.",
  title     = "UPAMNet: A unified network with deep knowledge priors for photoacoustic microscopy",
  journal   = "Photoacoustics",
  volume    = "38",
  pages     = "100608",
  year      = "2024",
  doi       = "10.1016/j.pacs.2024.100608"
}

@article{91he2022noising,
  author    = "He, D. and Zhou, J. and Shang, X. and Tang, X. and Luo, J. and Chen, S.-L.",
  title     = "De-noising of photoacoustic microscopy images by attentive generative adversarial network",
  journal   = "IEEE Trans. Med. Imaging",
  volume    = "42",
  number    = "5",
  pages     = "1349--1362",
  year      = "2022",
  doi       = "10.1109/TMI.2022.3224567"
}

@article{92zhang2023organ,
  author    = "Zhang, J. and Peng, D. and Qin, W. and Qi, W. and Liu, X. and Luo, Y. and Guo, Q. and Xi, L.",
  title     = "Organ-PAM: photoacoustic microscopy of whole-organ multiset vessel systems",
  journal   = "Laser Photonics Rev.",
  volume    = "17",
  number    = "7",
  pages     = "2201031",
  year      = "2023",
  doi       = "10.1002/lpor.202201031"
}

@article{93zhang2023adaptive,
  author    = "Zhang, Z. and Jin, H. and Zhang, W. and Lu, W. and Zheng, Z. and Sharma, A. and Pramanik, M. and Zheng, Y.",
  title     = "Adaptive enhancement of acoustic resolution photoacoustic microscopy imaging via deep CNN prior",
  journal   = "Photoacoustics",
  volume    = "30",
  pages     = "100484",
  year      = "2023",
  doi       = "10.1016/j.pacs.2023.100484"
}

@article{94meng2022depth,
  author    = "Meng, J. and Zhang, X. and Liu, L. and Zeng, S. and Fang, C. and Liu, C.",
  title     = "Depth-extended acoustic-resolution photoacoustic microscopy based on a two-stage deep learning network",
  journal   = "Biomed. Opt. Express",
  volume    = "13",
  number    = "8",
  pages     = "4386--4397",
  year      = "2022",
  doi       = "10.1364/BOE.461393"
}

@article{96li2024endosrr,
  author    = "Li, W. and Jia, F. and Liu, W.",
  title     = "EndoSRR: a comprehensive multi-stage approach for endoscopic specular reflection removal",
  journal   = "Int. J. Comput. Assist. Radiol. Surg.",
  volume    = "19",
  number    = "6",
  pages     = "1203--1211",
  year      = "2024",
  doi       = "10.1007/s11548-024-03138-5"
}

@article{97yeung2021focus,
  author    = "Yeung, M. and Sala, E. and Sch{\"o}nlieb, C.-B. and Rundo, L.",
  title     = "Focus U-Net: A novel dual attention-gated CNN for polyp segmentation during colonoscopy",
  journal   = "Comput. Biol. Med.",
  volume    = "137",
  pages     = "104815",
  year      = "2021",
  doi       = "10.1016/j.compbiomed.2021.104815"
}

@article{98gulenko2022deep,
  author    = "Gulenko, O. and Yang, H. and Kim, K. and Youm, J. Y. and Kim, M. and Kim, Y. and Jung, W. and Yang, J.-M.",
  title     = "Deep-learning-based algorithm for the removal of electromagnetic interference noise in photoacoustic endoscopic image processing",
  journal   = "Sensors",
  volume    = "22",
  number    = "10",
  pages     = "3961",
  year      = "2022",
  doi       = "10.3390/s22103961"
}

@article{99moghtaderi2024endoscopic,
  author    = "Moghtaderi, S. and Yaghoobian, O. and Wahid, K. A. and Lukong, K. E.",
  title     = "Endoscopic image enhancement: Wavelet transform and guided filter decomposition-based fusion approach",
  journal   = "J. Imaging",
  volume    = "10",
  number    = "1",
  pages     = "28",
  year      = "2024",
  doi       = "10.3390/jimaging10010028"
}

@article{100wang2022photoacoustic,
  author    = "Wang, Y. and Yuan, C. and Jiang, J. and Peng, K. and Wang, B.",
  title     = "Photoacoustic/ultrasound endoscopic imaging reconstruction algorithm based on the approximate Gaussian acoustic field",
  journal   = "Biosensors",
  volume    = "12",
  number    = "7",
  pages     = "463",
  year      = "2022",
  doi       = "10.3390/bios12070463"
}

@article{101xiao2022acoustic,
  author    = "Xiao, J. and Jiang, J. and Zhang, J. and Wang, Y. and Wang, B.",
  title     = "Acoustic-resolution-based spectroscopic photoacoustic endoscopy towards molecular imaging in deep tissues",
  journal   = "Opt. Express",
  volume    = "30",
  number    = "19",
  pages     = "35014--35028",
  year      = "2022",
  doi       = "10.1364/OE.467321"
}

@inproceedings{102yang2019quantitative,
  author    = "Yang, C. and Lan, H. and Zhong, H. and Gao, F.",
  title     = "Quantitative photoacoustic blood oxygenation imaging using deep residual and recurrent neural network",
  booktitle = "2019 {IEEE} 16th International Symposium on Biomedical Imaging ({ISBI})",
  pages     = "741--744",
  year      = "2019",
  address   = "Venice, Italy",
  publisher = "IEEE",
  doi       = "10.1109/ISBI.2019.8759223"
}

@article{103zhang2024navigating,
  author    = "Zhang, R. and A'dawiah, R. and Choo, T. W. J. and Li, X. and Balasundaram, G. and Qi, Y. and Goh, Y. and Bi, R. and Olivo, M.",
  title     = "Navigating challenges and solutions in quantitative photoacoustic imaging",
  journal   = "Appl. Phys. Rev.",
  volume    = "11",
  number    = "3",
  pages     = "031308",
  year      = "2024",
  doi       = "10.1063/5.0205587"
}

@article{104dantuma2021tunable,
  author    = "Dantuma, M. and Kruitwagen, S. and Ortega-Julia, J. and Pompe van Meerdervoort, R. P. and Manohar, S.",
  title     = "Tunable blood oxygenation in the vascular anatomy of a semi-anthropomorphic photoacoustic breast phantom",
  journal   = "J. Biomed. Opt.",
  volume    = "26",
  number    = "3",
  pages     = "036003",
  year      = "2021",
  doi       = "10.1117/1.JBO.26.3.036003"
}

@article{105zhou2020evaluation,
  author    = "Zhou, X. and Akhlaghi, N. and Wear, K. A. and Garra, B. S. and Pfefer, T. J. and Vogt, W. C.",
  title     = "Evaluation of fluence correction algorithms in multispectral photoacoustic imaging",
  journal   = "Photoacoustics",
  volume    = "19",
  pages     = "100181",
  year      = "2020",
  doi       = "10.1016/j.pacs.2020.100181"
}

@article{106zhao2017optical,
  author    = "Zhao, L. and Yang, M. and Jiang, Y. and Li, C.",
  title     = "Optical fluence compensation for handheld photoacoustic probe: An in vivo human study case",
  journal   = "J. Innov. Opt. Health Sci.",
  volume    = "10",
  number    = "04",
  pages     = "1740002",
  year      = "2017",
  doi       = "10.1142/S1793545817400023"
}

@article{107kirillin2017fluence,
  author    = "Kirillin, M. and Perekatova, V. and Turchin, I. and Subochev, P.",
  title     = "Fluence compensation in raster-scan optoacoustic angiography",
  journal   = "Photoacoustics",
  volume    = "8",
  pages     = "59--67",
  year      = "2017",
  doi       = "10.1016/j.pacs.2017.09.001"
}

@article{108hirasawa2016multispectral,
  author    = "Hirasawa, T. and Iwatate, R. J. and Kamiya, M. and Okawa, S. and Urano, Y. and Ishihara, M.",
  title     = "Multispectral photoacoustic imaging of tumours in mice injected with an enzyme-activatable photoacoustic probe",
  journal   = "J. Opt.",
  volume    = "19",
  number    = "1",
  pages     = "014002",
  year      = "2016",
  doi       = "10.1088/2040-8978/19/1/014002"
}

@article{109ranasinghesagara2014reflection,
  author    = "Ranasinghesagara, J. C. and Jiang, Y. and Zemp, R. J.",
  title     = "Reflection-mode multiple-illumination photoacoustic sensing to estimate optical properties",
  journal   = "Photoacoustics",
  volume    = "2",
  number    = "1",
  pages     = "33--38",
  year      = "2014",
  doi       = "10.1016/j.pacs.2014.01.001"
}

@article{110jin2019single,
  author    = "Jin, H. and Zhang, R. and Liu, S. and Zheng, Z. and Zheng, Y.",
  title     = "A single sensor dual-modality photoacoustic fusion imaging for compensation of light fluence variation",
  journal   = "IEEE Trans. Biomed. Eng.",
  volume    = "66",
  number    = "6",
  pages     = "1810--1813",
  year      = "2019",
  doi       = "10.1109/TBME.2018.2876532"
}

@article{112caravaca2013high,
  author    = "Caravaca-Aguirre, A. M. and Conkey, D. B. and Dove, J. D. and Ju, H. and Murray, T. W. and Piestun, R.",
  title     = "High contrast three-dimensional photoacoustic imaging through scattering media by localized optical fluence enhancement",
  journal   = "Opt. Express",
  volume    = "21",
  number    = "22",
  pages     = "26671--26676",
  year      = "2013",
  doi       = "10.1364/OE.21.026671"
}

@article{113fadhel2020fluence,
  author    = "Fadhel, M. N. and Hysi, E. and Assi, H. and Kolios, M. C.",
  title     = "Fluence-matching technique using photoacoustic radiofrequency spectra for improving estimates of oxygen saturation",
  journal   = "Photoacoustics",
  volume    = "19",
  pages     = "100182",
  year      = "2020",
  doi       = "10.1016/j.pacs.2020.100182"
}

@article{114dolet2021vitro,
  author    = "Dolet, A. and Ammanouil, R. and Petrilli, V. and Richard, C. and Tortoli, P. and Vray, D. and Varray, F.",
  title     = "In vitro and in vivo multispectral photoacoustic imaging for the evaluation of chromophore concentration",
  journal   = "Sensors",
  volume    = "21",
  number    = "10",
  pages     = "3366",
  year      = "2021",
  doi       = "10.3390/s21103366"
}

@article{115arabul2019unmixing,
  author    = "Arabul, M. U. and Rutten, M. C. M. and Bruneval, P. and van Sambeek, M. R. H. M. and van de Vosse, F. N. and Lopata, R. G. P.",
  title     = "Unmixing multi-spectral photoacoustic sources in human carotid plaques using non-negative independent component analysis",
  journal   = "Photoacoustics",
  volume    = "15",
  pages     = "100140",
  year      = "2019",
  doi       = "10.1016/j.pacs.2019.100140"
}

@article{116shi2014incorporating,
  author    = "Shi, C. and Wang, L.",
  title     = "Incorporating spatial information in spectral unmixing: A review",
  journal   = "Remote Sens. Environ.",
  volume    = "149",
  pages     = "70--87",
  year      = "2014",
  doi       = "10.1016/j.rse.2014.03.034"
}

@article{117tzoumas2017spectral,
  author    = "Tzoumas, S. and Ntziachristos, V.",
  title     = "Spectral unmixing techniques for optoacoustic imaging of tissue pathophysiology",
  journal   = "Philos. Trans. R. Soc. A",
  volume    = "375",
  number    = "2107",
  pages     = "20170262",
  year      = "2017",
  doi       = "10.1098/rsta.2017.0262"
}

@article{118feng2021detection,
  author    = "Feng, T. and Ge, Y. and Xie, Y. and Xie, W. and Liu, C. and Li, L. and Ta, D. and Jiang, Q. and Cheng, Q.",
  title     = "Detection of collagen by multi-wavelength photoacoustic analysis as a biomarker for bone health assessment",
  journal   = "Photoacoustics",
  volume    = "24",
  pages     = "100296",
  year      = "2021",
  doi       = "10.1016/j.pacs.2021.100296"
}

@article{119tzoumas2016eigenspectra,
  author    = "Tzoumas, S. and Nunes, A. and Olefir, I. and Stangl, S. and Symvoulidis, P. and Glasl, S. and Bayer, C. and Multhoff, G. and Ntziachristos, V.",
  title     = "Eigenspectra optoacoustic tomography achieves quantitative blood oxygenation imaging deep in tissues",
  journal   = "Nat. Commun.",
  volume    = "7",
  number    = "1",
  pages     = "12121",
  year      = "2016",
  doi       = "10.1038/ncomms12121"
}

@article{120karoui2012blind,
  author    = "Karoui, M. S. and Deville, Y. and Hosseini, S. and Ouamri, A.",
  title     = "Blind spatial unmixing of multispectral images: New methods combining sparse component analysis, clustering and non-negativity constraints",
  journal   = "Pattern Recognit.",
  volume    = "45",
  number    = "12",
  pages     = "4263--4278",
  year      = "2012",
  doi       = "10.1016/j.patcog.2012.05.010"
}

@article{121kirchner2018context,
  author    = "Kirchner, T. and Gr{\"o}hl, J. and Maier-Hein, L.",
  title     = "Context encoding enables machine learning-based quantitative photoacoustics",
  journal   = "J. Biomed. Opt.",
  volume    = "23",
  number    = "5",
  pages     = "056008",
  year      = "2018",
  doi       = "10.1117/1.JBO.23.5.056008"
}

@article{122cai2018end,
  author    = "Cai, C. and Deng, K. and Ma, C. and Luo, J.",
  title     = "End-to-end deep neural network for optical inversion in quantitative photoacoustic imaging",
  journal   = "Opt. Lett.",
  volume    = "43",
  number    = "12",
  pages     = "2752--2755",
  year      = "2018",
  doi       = "10.1364/OL.43.002752"
}

@article{123olefir2020deep,
  author  = "Olefir, I. and Tzoumas, S. and Restivo, C. and Mohajerani, P. and Xing, L. and Ntziachristos, V.",
  title   = "Deep learning-based spectral unmixing for optoacoustic imaging of tissue oxygen saturation",
  journal = "IEEE {T}rans. {M}ed. {I}maging",
  volume  = "39",
  number  = "11",
  pages   = "3643--3654",
  year    = "2020",
  doi     = "10.1109/TMI.2020.3003056"
}

@article{125dong2015image,
  author    = "Dong, C. and Loy, C. C. and He, K. and Tang, X.",
  title     = "Image super-resolution using deep convolutional networks",
  journal   = "IEEE Trans. Pattern Anal. Mach. Intell.",
  volume    = "38",
  number    = "2",
  pages     = "295--307",
  year      = "2015",
  doi       = "10.1109/TPAMI.2015.2439281"
}

@inproceedings{126dong2014learning,
  author    = "Dong, C. and Loy, C. C. and He, K. and Tang, X.",
  title     = "Learning a deep convolutional network for image super-resolution",
  booktitle = "European Conference on Computer Vision (ECCV)",
  series    = "Lecture Notes in Computer Science",
  volume    = "8692",
  pages     = "184--199",
  year      = "2014",
  publisher = "Springer",
  address   = "Cham",
  doi       = "10.1007/978-3-319-10590-1_13"
}

@inproceedings{127ledig2017photo,
  author    = "Ledig, C. and Theis, L. and Husz{\'a}r, F. and Caballero, J. and Cunningham, A. and Acosta, A. and Aitken, A. and Tejani, A. and Totz, J. and Wang, Z. and others",
  title     = "Photo-realistic single image super-resolution using a generative adversarial network",
  booktitle = "Proceedings of the {IEEE} Conference on Computer Vision and Pattern Recognition ({CVPR})",
  pages     = "4681--4690",
  year      = "2017",
  doi       = "10.1109/CVPR.2017.500"
}

@inproceedings{128wang2024camixersr,
  author    = "Wang, Y. and Liu, Y. and Zhao, S. and Li, J. and Zhang, L.",
  title     = "{CAMixerSR}: Only Details Need More Attention",
  booktitle = "Proceedings of the {IEEE}/{CVF} Conference on Computer Vision and Pattern Recognition ({CVPR})",
  pages     = "25837--25846",
  year      = "2024",
  doi       = "10.1109/CVPR52729.2024.02521"
}

@inproceedings{129sun2024coser,
  author    = "Sun, H. and Li, W. and Liu, J. and Chen, H. and Pei, R. and Zou, X. and Yan, Y. and Yang, Y.",
  title     = "{Coser}: Bridging image and language for cognitive super-resolution",
  booktitle = "Proceedings of the IEEE/CVF Conference on Computer Vision and Pattern Recognition (CVPR)",
  pages     = "25868--25878",
  year      = "2024",
  doi       = "10.1109/CVPR52729.2024.02524"
}

@inproceedings{130kim2016accurate,
  author    = "Kim, J. and Lee, J. K. and Lee, K. M.",
  title     = "Accurate image super-resolution using very deep convolutional networks",
  booktitle = "Proceedings of the IEEE Conference on Computer Vision and Pattern Recognition (CVPR)",
  pages     = "1646--1654",
  year      = "2016",
  doi       = "10.1109/CVPR.2016.182"
}

@inproceedings{131tai2017image,
  author    = "Tai, Y. and Yang, J. and Liu, X.",
  title     = "Image super-resolution via deep recursive residual network",
  booktitle = "Proceedings of the IEEE Conference on Computer Vision and Pattern Recognition (CVPR)",
  pages     = "3147--3155",
  year      = "2017",
  doi       = "10.1109/CVPR.2017.335"
}

@inproceedings{132liu2020residual,
  author    = "Liu, J. and Tang, J. and Wu, G.",
  title     = "Residual feature distillation network for lightweight image super-resolution",
  booktitle = "European Conference on Computer Vision (ECCV)",
  series    = "Lecture Notes in Computer Science",
  volume    = "12365",
  pages     = "41--55",
  year      = "2020",
  publisher = "Springer",
  address   = "Cham, Switzerland",
  doi       = "10.1007/978-3-030-58545-7_3"
}

@inproceedings{133wang2019transform,
  author    = "Wang, C. and Wang, S. and Ma, B. and Li, J. and Dong, X. and Xia, Z.",
  title     = "Transform domain based medical image super-resolution via deep multi-scale network",
  booktitle = "2019 {IEEE} International Conference on Acoustics, Speech and Signal Processing ({ICASSP})",
  pages     = "2387--2391",
  year      = "2019",
  address   = "Brighton, UK",
  publisher = "IEEE",
  doi       = "10.1109/ICASSP.2019.8683388"
}

@article{134shen2024physics,
  author  = "Shen, K. and Niu, K. and Liu, S. and Paulus, Y. M. and Jiang, X. and Tian, C.",
  title   = "Physics-driven deep learning photoacoustic tomography",
  journal = "Fundamental Research",
  year    = "2025",
  note    = "In press",
  doi     = "10.1016/j.fmre.2024.12.005"
}

@article{135zhu2023novel,
  author    = "Zhu, F. and Liu, W.",
  title     = "A novel medical image fusion method based on multi-scale shearing rolling weighted guided image filter",
  journal   = "Math. Biosci. Eng.",
  volume    = "20",
  number    = "8",
  pages     = "15374--15406",
  year      = "2023",
  doi       = "10.3934/mbe.2023678"
}

@article{136ma2020ganmcc,
  author    = "Ma, J. and Zhang, H. and Shao, Z. and Liang, P. and Xu, H.",
  title     = "{GANMcC}: A generative adversarial network with multiclassification constraints for infrared and visible image fusion",
  journal   = "IEEE Trans. Instrum. Meas.",
  volume    = "70",
  pages     = "1--14",
  year      = "2020",
  doi       = "10.1109/TIM.2020.3028324"
}

@article{137nyayapathi2024dual,
  author    = "Nyayapathi, N. and Zheng, E. and Zhou, Q. and Doyley, M. and Xia, J.",
  title     = "Dual-modal photoacoustic and ultrasound imaging: from preclinical to clinical applications",
  journal   = "Front. Photonics",
  volume    = "5",
  pages     = "1359784",
  year      = "2024",
  doi       = "10.3389/fphoton.2024.1359784"
}

@article{138notsuka2022improvement,
  author    = "Notsuka, Y. and Kurihara, M. and Hashimoto, N. and Harada, Y. and Takahashi, E. and Yamaoka, Y.",
  title     = "Improvement of spatial resolution in photoacoustic microscopy using transmissive adaptive optics with a low-frequency ultrasound transducer",
  journal   = "Opt. Express",
  volume    = "30",
  number    = "2",
  pages     = "2933--2948",
  year      = "2022",
  doi       = "10.1364/OE.447212"
}

@article{139biswas2016method,
  author    = "Biswas, S. K. and van Es, P. and Steenbergen, W. and Manohar, S.",
  title     = "A method for delineation of bone surfaces in photoacoustic computed tomography of the finger",
  journal   = "Ultrason. Imaging",
  volume    = "38",
  number    = "1",
  pages     = "63--76",
  year      = "2016",
  doi       = "10.1177/0161734615583072"
}

@article{140le2022segmentation,
  author  = "Le, T. D. and Kwon, S.-Y. and Lee, C.",
  title   = "Segmentation and quantitative analysis of photoacoustic imaging: a review",
  journal = "Photonics",
  volume  = "9",
  number  = "3",
  pages   = "176",
  year    = "2022",
  doi     = "10.3390/photonics9030176"
}

@article{141liang2022optical,
  author    = "Liang, Y. and Fu, W. and Li, Q. and Chen, X. and Sun, H. and Wang, L. and Jin, L. and Huang, W. and Guan, B.-O.",
  title     = "Optical-resolution functional gastrointestinal photoacoustic endoscopy based on optical heterodyne detection of ultrasound",
  journal   = "Nat. Commun.",
  volume    = "13",
  number    = "1",
  pages     = "7604",
  year      = "2022",
  doi       = "10.1038/s41467-022-35386-9"
}

@article{142sun2025dual,
  author  = "Sun, H. and Liang, X. and Li, L. and Zhao, Y. and Guo, H. and Qi, W. and Xi, L.",
  title   = "Dual-Scanning Photoacoustic Endomicroscopy for High-Speed Gastrointestinal Microvascular Imaging",
  journal = "{IEEE} Transactions on Medical Imaging",
  year    = "2025",
  note    = "Early Access",
  doi     = "10.1109/TMI.2025.3541234"
}

@article{143zhang2025recent,
  author    = "Zhang, K. and Ge, N. and Shen, L. and Yang, F. and Qiu, J. and Guo, J. and Wang, K. and Wang, S. and Yang, F. and Sheng, S. and others",
  title     = "Recent research progress of photoacoustic endoscopy in the digestive system",
  journal   = "Endosc. Ultrasound",
  volume    = "14",
  number    = "3",
  pages     = "99--105",
  year      = "2025",
  doi       = "10.4103/EUS-D-24-00087"
}

@article{144yang2024perspectives,
  author    = "Yang, S. and Hu, S.",
  title     = "Perspectives on endoscopic functional photoacoustic microscopy",
  journal   = "Appl. Phys. Lett.",
  volume    = "125",
  number    = "3",
  pages     = "030501",
  year      = "2024",
  doi       = "10.1063/5.0215678"
}

@article{145shang2017simultaneous,
  author    = "Shang, S. and Chen, Z. and Zhao, Y. and Yang, S. and Xing, D.",
  title     = "Simultaneous imaging of atherosclerotic plaque composition and structure with dual-mode photoacoustic and optical coherence tomography",
  journal   = "Opt. Express",
  volume    = "25",
  number    = "2",
  pages     = "530--539",
  year      = "2017",
  doi       = "10.1364/OE.25.000530"
}

@article{146xu2006photoacoustic,
  author    = "Xu, M. and Wang, L. V.",
  title     = "Photoacoustic imaging in biomedicine",
  journal   = "Rev. Sci. Instrum.",
  volume    = "77",
  number    = "4",
  pages     = "041101",
  year      = "2006",
  doi       = "10.1063/1.2195024"
}

@article{147deng2025streamlined,
  author    = "Deng, H. and Zhou, Y. and Xiang, J. and Gu, L. and Luo, Y. and Feng, H. and Liu, M. and Ma, C.",
  title     = "Streamlined photoacoustic image processing with foundation models: A training-free solution",
  journal   = "J. Innov. Opt. Health Sci.",
  volume    = "18",
  number    = "01",
  pages     = "2450019",
  year      = "2025",
  doi       = "10.1142/S1793545824500198"
}

@article{148lan2024masked,
  author    = "Lan, H. and Huang, L. and Wei, X. and Li, Z. and Lv, J. and Ma, C. and Nie, L. and Luo, J.",
  title     = "Masked cross-domain self-supervised deep learning framework for photoacoustic computed tomography reconstruction",
  journal   = "Neural Netw.",
  volume    = "179",
  pages     = "106515",
  year      = "2024",
  doi       = "10.1016/j.neunet.2024.106515"
}

@article{149song2023sparse,
  author    = "Song, X. and Wang, G. and Zhong, W. and Guo, K. and Li, Z. and Liu, X. and Dong, J. and Liu, Q.",
  title     = "Sparse-view reconstruction for photoacoustic tomography combining diffusion model with model-based iteration",
  journal   = "Photoacoustics",
  volume    = "33",
  pages     = "100558",
  year      = "2023",
  doi       = "10.1016/j.pacs.2023.100558"
}

@article{150song2024multiple,
  author    = "Song, X. and Zou, X. and Zeng, K. and Li, J. and Hou, S. and Wu, Y. and Li, Z. and Ma, C. and Zheng, Z. and Guo, K. and others",
  title     = "Multiple diffusion models-enhanced extremely limited-view reconstruction strategy for photoacoustic tomography boosted by multi-scale priors",
  journal   = "Photoacoustics",
  volume    = "40",
  pages     = "100646",
  year      = "2024",
  doi       = "10.1016/j.pacs.2024.100646"
}

@article{151lian2025generative,
  author    = "Lian, T. and Lv, Y. and Guo, K. and Li, Z. and Li, J. and Wang, G. and Lin, J. and Cao, Y. and Liu, Q. and Song, X.",
  title     = "Generative priors-constraint accelerated iterative reconstruction for extremely sparse photoacoustic tomography boosted by mean-reverting diffusion model: Towards 8 projections",
  journal   = "Photoacoustics",
  volume    = "43",
  pages     = "100709",
  year      = "2025",
  doi       = "10.1016/j.pacs.2025.100709"
}

@article{152wang2012photoacoustic,
  author    = "Wang, L. V. and Hu, S.",
  title     = "Photoacoustic tomography: in vivo imaging from organelles to organs",
  journal   = "Science",
  volume    = "335",
  number    = "6075",
  pages     = "1458--1462",
  year      = "2012",
  doi       = "10.1126/science.1216210"
}

@article{153yin2021organic,
  author    = "Yin, C. and Li, X. and Wang, Y. and Liang, Y. and Zhou, S. and Zhao, P. and Lee, C.-S. and Fan, Q. and Huang, W.",
  title     = "Organic semiconducting macromolecular dyes for NIR-II photoacoustic imaging and photothermal therapy",
  journal   = "Adv. Funct. Mater.",
  volume    = "31",
  number    = "37",
  pages     = "2104650",
  year      = "2021",
  doi       = "10.1002/adfm.202104650"
}

@article{154gao2025nir,
  author    = "Gao, B. and Gao, L. and Wang, J. and Huang, X. and Ge, W. and Li, S. and Wang, F.",
  title     = "{NIR-II} Light Field Microscopy for Through-Skull Hemodynamic Volumetric Imaging in Awake Mice",
  journal   = "Laser Photonics Rev.",
  pages     = "2401685",
  year      = "2025",
  note      = "Early view",
  doi       = "10.1002/lpor.202401685"
}

@inproceedings{155dasa2019towards,
  author    = "Dasa, M. K. and Markos, C. and Janting, J. and Adamu, A. I. and Bowen, P. and Bang, O.",
  title     = "Towards accurate and label-free monitoring of bio-analytes using supercontinuum based multispectral photoacoustic spectroscopy in the extended near-infrared wavelength regime",
  booktitle = "Label-free Biomedical Imaging and Sensing ({LBIS}) 2019",
  series    = "Proceedings of {SPIE}",
  volume    = "10890",
  pages     = "108900N",
  year      = "2019",
  address   = "San Francisco, California, United States",
  publisher = "SPIE",
  doi       = "10.1117/12.2509050"
}

@article{156shaked2023label,
  author    = "Shaked, N. T. and Boppart, S. A. and Wang, L. V. and Popp, J.",
  title     = "Label-free biomedical optical imaging",
  journal   = "Nat. Photonics",
  volume    = "17",
  number    = "12",
  pages     = "1031--1041",
  year      = "2023",
  doi       = "10.1038/s41566-023-01318-3"
}

@article{157dasa2020all,
  author    = "Dasa, M. K. and Nteroli, G. and Bowen, P. and Messa, G. and Feng, Y. and Petersen, C. R. and Koutsikou, S. and Bondu, M. and Moselund, P. M. and Podoleanu, A. and others",
  title     = "All-fibre supercontinuum laser for in vivo multispectral photoacoustic microscopy of lipids in the extended near-infrared region",
  journal   = "Photoacoustics",
  volume    = "18",
  pages     = "100163",
  year      = "2020",
  doi       = "10.1016/j.pacs.2020.100163"
}

@article{158zhu2020towards,
  author    = "Zhu, Y. and Feng, T. and Cheng, Q. and Wang, X. and Du, S. and Sato, N. and Yuan, J. and Singh, M. K. A.",
  title     = "Towards clinical translation of LED-based photoacoustic imaging: a review",
  journal   = "Sensors",
  volume    = "20",
  number    = "9",
  pages     = "2484",
  year      = "2020",
  doi       = "10.3390/s20092484"
}

@article{159hauptmann2018model,
  author    = "Hauptmann, A. and Lucka, F. and Betcke, M. and Huynh, N. and Adler, J. and Cox, B. and Beard, P. and Ourselin, S. and Arridge, S.",
  title     = "Model-based learning for accelerated, limited-view 3-D photoacoustic tomography",
  journal   = "IEEE Trans. Med. Imaging",
  volume    = "37",
  number    = "6",
  pages     = "1382--1393",
  year      = "2018",
  doi       = "10.1109/TMI.2018.2812760"
}

@article{160deng2021deep,
  author    = "Deng, H. and Qiao, H. and Dai, Q. and Ma, C.",
  title     = "Deep learning in photoacoustic imaging: a review",
  journal   = "J. Biomed. Opt.",
  volume    = "26",
  number    = "4",
  pages     = "040901",
  year      = "2021",
  doi       = "10.1117/1.JBO.26.4.040901"
}

@article{161murray2016super,
  author    = "Murray, T. W. and Haltmeier, M. and Berer, T. and Leiss-Holzinger, E. and Burgholzer, P.",
  title     = "Super-resolution photoacoustic microscopy using blind structured illumination",
  journal   = "Optica",
  volume    = "4",
  number    = "1",
  pages     = "17--22",
  year      = "2016",
  doi       = "10.1364/OPTICA.4.000017"
}

@article{162manohar2011gold,
  author    = "Manohar, S. and Ungureanu, C. and Van Leeuwen, T. G.",
  title     = "Gold nanorods as molecular contrast agents in photoacoustic imaging: the promises and the caveats",
  journal   = "Contrast Media Mol. Imaging",
  volume    = "6",
  number    = "5",
  pages     = "389--400",
  year      = "2011",
  doi       = "10.1002/cmmi.445"
}

@article{163zhang2024exploiting,
  author    = "Zhang, J. and Zhu, Y.",
  title     = "Exploiting the Photo-Physical Properties of Metal Halide Perovskite Nanocrystals for Bioimaging",
  journal   = "ChemBioChem",
  volume    = "25",
  number    = "5",
  pages     = "e202300683",
  year      = "2024",
  doi       = "10.1002/cbic.202300683"
}

@article{164hsu2023nanomaterial,
  author    = "Hsu, J. C. and Tang, Z. and Eremina, O. E. and Sofias, A. M. and Lammers, T. and Lovell, J. F. and Zavaleta, C. and Cai, W. and Cormode, D. P.",
  title     = "Nanomaterial-based contrast agents",
  journal   = "Nat. Rev. Methods Primers",
  volume    = "3",
  number    = "1",
  pages     = "30",
  year      = "2023",
  doi       = "10.1038/s43586-023-00205-8"
}

@article{165lee2023panoramic,
  author    = "Lee, C. and Cho, S. and Lee, D. and Lee, J. and Park, J.-I. and Kim, H.-J. and Park, S. H. and Choi, W. and Kim, U. and Kim, C.",
  title     = "Panoramic volumetric clinical handheld photoacoustic and ultrasound imaging",
  journal   = "Photoacoustics",
  volume    = "31",
  pages     = "100512",
  year      = "2023",
  doi       = "10.1016/j.pacs.2023.100512"
}

@article{166zheng2025physics,
  author    = "Zheng, S. and Zhu, A. and Hou, Y. and Sun, M.",
  title     = "A physics-based iterative learning framework for quantitative parametric imaging with application to photoacoustic imaging",
  journal   = "Eng. Appl. Artif. Intell.",
  volume    = "142",
  pages     = "109920",
  year      = "2025",
  doi       = "10.1016/j.engappai.2025.109920"
}

@article{167liang2024deep,
  author    = "Liang, Z. and Zhang, S. and Liang, Z. and Mo, Z. and Zhang, X. and Zhong, Y. and Chen, W. and Qi, L.",
  title     = "Deep learning acceleration of iterative model-based light fluence correction for photoacoustic tomography",
  journal   = "Photoacoustics",
  volume    = "37",
  pages     = "100601",
  year      = "2024",
  doi       = "10.1016/j.pacs.2024.100601"
}

@article{168antun2020instabilities,
  author    = "Antun, V. and Renna, F. and Poon, C. and Adcock, B. and Hansen, A. C.",
  title     = "On instabilities of deep learning in image reconstruction and the potential costs of AI",
  journal   = "Proc. Natl. Acad. Sci. USA",
  volume    = "117",
  number    = "48",
  pages     = "30088--30095",
  year      = "2020",
  doi       = "10.1073/pnas.1914015117"
}

@article{169yang2023deep,
  author    = "Yang, D. and Ran, A. R. and Nguyen, T. X. and Lin, T. P. H. and Chen, H. and Lai, T. Y. Y. and Tham, C. C. and Cheung, C. Y.",
  title     = "Deep learning in optical coherence tomography angiography: Current progress, challenges, and future directions",
  journal   = "Diagnostics",
  volume    = "13",
  number    = "2",
  pages     = "326",
  year      = "2023",
  doi       = "10.3390/diagnostics13020326"
}

@book{170montavon2019explainable,
  editor    = "Montavon, G. and Binder, A. and Lapuschkin, S. and Samek, W. and M{\"u}ller, K.-R.",
  title     = "Explainable {AI}: Interpreting, Explaining and Visualizing Deep Learning",
  series    = "Lecture Notes in Computer Science",
  volume    = "11700",
  year      = "2019",
  publisher = "Springer",
  address   = "Cham, Switzerland",
  doi       = "10.1007/978-3-030-28954-6"
}

@article{171hauptmann2020deep,
  author    = "Hauptmann, A. and Cox, B.",
  title     = "Deep learning in photoacoustic tomography: current approaches and future directions",
  journal   = "J. Biomed. Opt.",
  volume    = "25",
  number    = "11",
  pages     = "112903",
  year      = "2020",
  doi       = "10.1117/1.JBO.25.11.112903"
}

@inproceedings{172kendall2017uncertainties,
  author    = "Kendall, A. and Gal, Y.",
  title     = "What uncertainties do we need in {B}ayesian deep learning for computer vision?",
  booktitle = "Advances in Neural Information Processing Systems (NeurIPS)",
  volume    = "30",
  pages     = "5575--5585",
  year      = "2017",
  publisher = "Curran Associates, Inc.",
  address   = "Red Hook, NY",
}

@article{173langley2025heterogeneous,
  author    = "Langley, A. and Sweeney, A. and Shethia, R. T. and Bednarke, B. and Wulandana, F. and Xavierselvan, M. and Mallidi, S.",
  title     = "Heterogeneous tumor blood oxygenation dynamics during phototherapy deciphered with real-time label-free photoacoustic imaging",
  journal   = "npj Acoust.",
  volume    = "1",
  number    = "1",
  pages     = "9",
  year      = "2025",
  doi       = "10.1038/s44324-025-00009-8"
}

@article{174zhu2022real,
  author    = "Zhu, X. and Huang, Q. and DiSpirito, A. and Vu, T. and Rong, Q. and Peng, X. and Sheng, H. and Shen, X. and Zhou, Q. and Jiang, L. and others",
  title     = "Real-time whole-brain imaging of hemodynamics and oxygenation at micro-vessel resolution with ultrafast wide-field photoacoustic microscopy",
  journal   = "Light Sci. Appl.",
  volume    = "11",
  number    = "1",
  pages     = "138",
  year      = "2022",
  doi       = "10.1038/s41377-022-00828-9"
}

@article{175jiang2018estimation,
  author    = "Jiang, Y. and Zemp, R.",
  title     = "Estimation of cerebral metabolic rate of oxygen consumption using combined multiwavelength photoacoustic microscopy and Doppler microultrasound",
  journal   = "J. Biomed. Opt.",
  volume    = "23",
  number    = "1",
  pages     = "016009",
  year      = "2018",
  doi       = "10.1117/1.JBO.23.1.016009"
}

@article{176menozzi2025light,
  author    = "Menozzi, L. and Li, Z. and Choi, S. and Vu, T. and Shi, L. and Yao, J.",
  title     = "Light and metabolism: label-free optical imaging of metabolic activities in biological systems",
  journal   = "Biomed. Opt. Express",
  volume    = "16",
  number    = "9",
  pages     = "3770--3796",
  year      = "2025",
  doi       = "10.1364/BOE.558765"
}

@article{177wang2012vivo,
  author    = "Wang, B. and Karpiouk, A. and Yeager, D. and Amirian, J. and Litovsky, S. and Smalling, R. and Emelianov, S.",
  title     = "In vivo intravascular ultrasound-guided photoacoustic imaging of lipid in plaques using an animal model of atherosclerosis",
  journal   = "Ultrasound Med. Biol.",
  volume    = "38",
  number    = "12",
  pages     = "2098--2103",
  year      = "2012",
  doi       = "10.1016/j.ultrasmedbio.2012.07.015"
}

@article{178zhou2023multimodal,
  author    = "Zhou, W. and Chen, D. and Li, K. and Yuan, Z. and Chen, X.",
  title     = "Multimodal photoacoustic imaging in analytic vulnerability of atherosclerosis",
  journal   = "iRadiology",
  volume    = "1",
  number    = "4",
  pages     = "303--319",
  year      = "2023",
  doi       = "10.1002/ird3.23"
}

@article{179wang2016practical,
  author    = "Wang, L. V. and Yao, J.",
  title     = "A practical guide to photoacoustic tomography in the life sciences",
  journal   = "Nat. Methods",
  volume    = "13",
  number    = "8",
  pages     = "627--638",
  year      = "2016",
  doi       = "10.1038/nmeth.3914"
}

@book{180szabo2013diagnostic,
  author    = "Szabo, T. L.",
  title     = "Diagnostic Ultrasound Imaging: Inside Out",
  edition   = "2nd",
  year      = "2014",
  publisher = "Academic Press",
  address   = "Waltham, MA, USA",
  note      = "First edition 2004; second edition 2014"
}

@article{181zhao2023hybrid,
  author    = "Zhao, S. and Hartanto, J. and Joseph, R. and Wu, C.-H. and Zhao, Y. and Chen, Y.-S.",
  title     = "Hybrid photoacoustic and fast super-resolution ultrasound imaging",
  journal   = "Nat. Commun.",
  volume    = "14",
  number    = "1",
  pages     = "2191",
  year      = "2023",
  doi       = "10.1038/s41467-023-37885-3"
}

\end{document}